\documentclass[12pt]{article}

\usepackage[margin=0.5in]{geometry}

\usepackage[T1]{fontenc}

\usepackage{times}
\usepackage{xspace}
\usepackage[dvipsnames]{xcolor}
\usepackage{graphicx}
\usepackage{amsmath,amssymb}
\usepackage{booktabs}
\usepackage[numbers,sort&compress]{natbib}
\usepackage[format=plain,labelformat=simple]{caption}
\usepackage{subcaption}
\usepackage[hyphens]{url}
\usepackage{authblk}

\usepackage[breaklinks,colorlinks,allcolors=blue]{hyperref}
\usepackage{cleveref}

\usepackage{multirow}

\title{Continual Visual Anomaly Detection on the Edge:\\ Benchmark and Efficient Solutions}

\author[1]{Manuel Barusco}
\author[2]{Francesco Borsatti}
\author[3]{David Petrovic}
\author[4]{Davide Dalle Pezze}
\author[5]{Gian Antonio Susto}
\affil[ ]{\centering\normalsize University of Padova, Italy}

\affil[1]{\texttt{manuel.barusco@phd.unipd.it}}
\affil[2]{\texttt{francesco.borsatti.1@phd.unipd.it}}
\affil[3]{\texttt{david.petrovic@studenti.unipd.it}}
\affil[4]{\texttt{davide.dallepezze@unipd.it}}
\affil[5]{\texttt{gianantonio.susto@unipd.it}}
\date{}

\begin{document}
\maketitle

\begin{abstract}
Visual Anomaly Detection (VAD) is a critical task for many applications including industrial inspection and healthcare.
While VAD has been extensively studied, two key challenges remain largely unaddressed in conjunction: edge deployment, where computational resources are severely constrained, and continual learning, where models must adapt to evolving data distributions without forgetting previously acquired knowledge.
Our benchmark provides guidance for the selection of the optimal backbone and VAD method under joint efficiency and adaptability constraints, characterizing the trade-offs between memory footprint, inference cost, and detection performance.
Studying these challenges in isolation is insufficient, as methods designed for one setting make assumptions that break down when the other constraint is simultaneously imposed.
In this work, we propose the first comprehensive benchmark for VAD on the edge in the continual learning scenario, evaluating seven VAD models across three lightweight backbone architectures.
Furthermore, we propose Tiny-Dinomaly, a lightweight adaptation of the Dinomaly model built on the DINO foundation model that achieves a 11$\times$ smaller memory footprint and 20$\times$ lower computational cost while improving localization (Pixel F1) by 5 percentage points.
Finally, we introduce targeted modifications to PatchCore and PaDiM to improve their efficiency in the continual learning setting.
\end{abstract}


\section{Introduction}
\label{sec:introduction}

A critical computer vision task is Visual Anomaly Detection (VAD), which involves finding anomalous images and localizing the responsible pixels.
By leveraging unsupervised learning, VAD eliminates the need for costly pixel-level annotations, making it broadly applicable across industrial inspection \cite{huang2024prototype} and healthcare \cite{bao2024bmad}.

In recent years, VAD has drawn a lot of research interest. 
Nevertheless, the majority of current approaches were created and assessed under idealized assumptions, such as centralized server-side deployment, abundant computational resources, and static data distributions.

However, these requirements are rarely met by real-world deployments. Over time, new categories of objects are constantly introduced in industrial settings. This leads to shifts in the data distribution that a static model cannot handle.
Therefore, recent methods have spanned a range of approaches, from replay-based \cite{bugarin2024unveiling} to prompt-based \cite{liu2024unsupervised} techniques.
These methods show that incremental updates to VAD models are possible without catastrophic forgetting.
At the same time, VAD models are being pushed toward edge deployment by the increasing need for low-latency, energy-efficient, and privacy-preserving systems. 
In this setting, memory and processing power are severely limited.
According to recent studies, feature-based VAD techniques can be easily made edge-compatible by replacing standard backbones with lightweight alternatives \cite{barusco2025paste} and the detection performance is only slightly reduced as a result of this modification.

These two challenges, continual learning and edge deployment, have so far been studied largely in isolation.
Continual learning methods are typically benchmarked on unconstrained hardware. In this setting, large replay buffers and expensive update steps are affordable. However, when transferred to edge devices, these assumptions break down. Conversely, VAD models optimized for edge inference are designed for a fixed data distribution, without accounting for the additional memory and computational overhead that continual learning introduces. 
Only by addressing both challenges jointly can VAD systems become efficient, adaptive, and deployable on commodity devices. 
This is precisely the setting required by most industrial and real-world applications.

Only very recently has a first attempt been made to address both challenges jointly \cite{barusco2025memory}.
However, a comprehensive and systematic study of VAD in this combined setting is still missing. It remains unclear which VAD architectures are best suited for edge-constrained continual learning. 
It is also unclear which backbones offer the best trade-off between efficiency and performance. 
Finally, it is still an open question whether foundation-model-based approaches can be successfully adapted to this challenging regime.

To fill this gap, we propose the first comprehensive benchmark for continual VAD on the edge. We evaluate seven VAD models across three lightweight backbone architectures, providing a thorough analysis of their behavior in the continual edge scenario. In addition, we propose Tiny-Dinomaly, a lightweight adaptation of the Dinomaly model built on the DINO foundation model \cite{oquab2023dinov2}. The proposed method retains strong anomaly detection performance while meeting edge constraints. Finally, we introduce targeted modifications to PatchCore and PaDiM to improve their efficiency and suitability for continual learning on edge devices. We evaluate our benchmark on two widely adopted datasets, 
MVTec AD \citep{bergmann2019mvtec} and VisA \citep{zou2022spot},  which together cover a broad range of industrial inspection scenarios.

The main contributions of this work are as follows:
\begin{enumerate}
    \item We propose the first comprehensive benchmark for VAD models targeting edge deployment in the continual learning scenario
    \item We propose Tiny-Dinomaly, an edge-adapted version of Dinomaly built on the DINO foundation model, representing a strong candidate for efficient edge deployment in continual VAD settings.
    \item We propose targeted modifications to PaDiM and PatchCore to improve their efficiency and suitability for edge-constrained continual learning scenarios.
\end{enumerate}

The remainder of this paper is organized as follows. 
Section \ref{sec:related_work} reviews related work on Visual Anomaly Detection, continual learning, and edge deployment. 
Section \ref{sec:methodology} describes the methodology, including the proposed Tiny-Dinomaly architecture and the modifications to the original PaDiM-CL achieves anand PaDiM. 
Section \ref{sec:exp_settings} details the experimental settings, covering datasets, evaluation metrics, and model configurations. 
Section \ref{sec:results} presents and discusses the benchmark results, analyzing the impact of backbone choice and the performance of our proposed methods. 
Finally, Section \ref{sec:conclusion} concludes the paper and outlines directions for future work.

\begin{figure}[!thbp]
    \centering
    \includegraphics[width=0.8\textwidth]{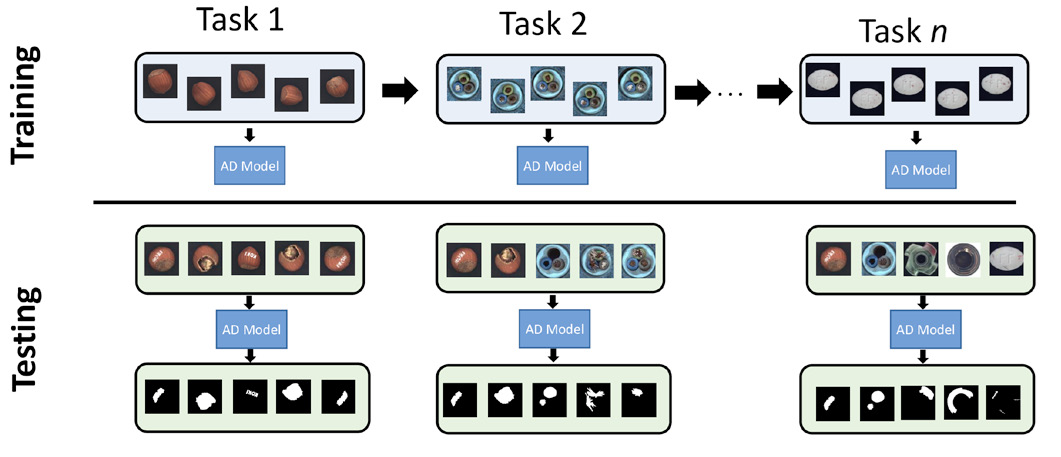}
    \caption{ Continual Learning setting for the VAD. Each task introduces a new item, the VAD model must identify anomalous products at the image level and localize defects at the pixel level for newly introduced items, while maintaining strong performance on previously learned categories.
    }
    \label{fig:cl_vad_scenario}
\end{figure}

\section{Related Work}
\label{sec:related_work}

In this section, we review the literature most relevant to our work. 
We first discuss recent advances in Visual Anomaly Detection, with a focus on feature-based approaches and their adaptation for edge deployment. 
We then survey continual learning methods for VAD and highlight the key limitations that motivate our benchmark and proposed solutions.

\subsection{Visual Anomaly Detection}
Visual Anomaly Detection (VAD) is a well-established Computer Vision task with broad applications in industrial inspection and healthcare. The goal is to identify anomalous images and localize the responsible pixels. This is achieved by leveraging unsupervised learning, which avoids the need for costly pixel-level annotations. Over the years, a large body of methods has been developed. These methods can be broadly categorized into reconstruction-based and feature-based approaches.

Reconstruction-based methods rely on generative models, such as autoencoders, GANs, and diffusion models, trained exclusively on normal images \cite{zavrtanik2021draem,zavrtanik2021reconstruction,he2024diffusion,zhang2025diffusionad}. At inference time, anomalies are detected as regions where the model fails to reconstruct the input faithfully. 
This results in large reconstruction errors. While conceptually simple, these approaches tend to be computationally expensive and can sometimes reconstruct anomalous regions accurately, limiting their detection reliability.

Feature-based approaches have emerged as the dominant paradigm in recent years, offering a better trade-off between performance and efficiency. These methods leverage pretrained neural networks as fixed feature extractors and build normality representations from the extracted embeddings. 
They can be further divided into three families. Memory-bank methods, such as PatchCore and PaDiM, store representative patch-level embeddings of normal images \cite{roth2022towards,defard2021padim,9839549}. Anomalies are detected by measuring the distance between test features and the stored representations. Teacher-student methods, such as STFPM, employ a frozen pretrained teacher and a trainable student network \cite{wang2021studentteacherfeaturepyramidmatching,zhang2023destseg,Deng_2022_CVPR}. Since the student is trained only on normal images, discrepancies between student and teacher feature maps at inference time indicate anomalies. Normalizing flow methods, such as FastFlow, model the distribution of normal features explicitly and use the estimated likelihood as an anomaly score \cite{yu2021fastflow,gudovskiy2022cflow}. 

Some recent approaches considered edge deployment for VAD models, for example \cite{barusco2025paste} evaluates several feature-based VAD methods after replacing their original backbones with lightweight alternatives. Building on this direction, subsequent works have proposed efficiency-oriented variants of established methods, including PaDiM-Lite \cite{stropeni2026efficient} and a lightweight version of STFPM introduced in \cite{barusco2025paste} called paSTe.

\subsection{Continual Learning for Edge VAD}
Continual Learning (CL) for VAD assumes that object categories are introduced sequentially as a stream of tasks, requiring the model to adapt to new data while preserving prior knowledge. This setting inherits the challenges of multi-class learning with the added difficulty of an evolving data distribution and the impossibility of accessing previous data when training on a new category, a constraint that makes catastrophic forgetting a central concern.

Early work applied standard CL strategies such as Replay to VAD \cite{bugarin2024unveiling}. Replay mitigates catastrophic forgetting by interleaving past and new data during training using a small replay buffer, where a limited fraction of previously seen samples is retained and reused across task boundaries \cite{chaudhry2019continual}. 

Building on this foundation, researchers have since proposed methods specifically designed to tackle VAD within the CL paradigm \cite{li2022towards,liu2024unsupervised,li2025one}.

A key motivation for CL is its efficiency; incremental updates to a model are significantly less computationally demanding than full retraining from scratch. By eliminating the need to re-process the entire historical dataset for every update, CL reduces the reliance on powerful remote servers and high-bandwidth cloud offloading. This shift makes it increasingly feasible to perform training and adaptation directly on local edge devices, ensuring data privacy and enabling real-time responsiveness in evolving environments.

However, achieving practical deployability on commodity devices requires pairing continual learning strategies with architectures specifically designed for edge constraints.
A recent work took a first step in this direction by evaluating STFPM, and PaSTe in a continual edge setting \cite{barusco2025memory}. Yet, this study remains limited in scope and falls short of a thorough analysis of the problem. They analyzed only two VAD methods on the edge (both of the student-teacher category), making it impossible to generalize the insights to different categories like memory bank approaches. 
Furthermore, the backbone architectures considered are limited in number, leaving open the question of which lightweight backbone offers the best efficiency-performance trade-off.

To address this gap, we propose the first comprehensive benchmark for continual VAD on the edge, covering seven VAD models across three lightweight backbone architectures. Our benchmark provides a detailed analysis of model behavior in the continual edge scenario. It also identifies which architectural choices are most suitable. Furthermore, we introduce Tiny-Dinomaly, a lightweight adaptation of Dinomaly that retains strong detection performance, and propose modifications to PatchCore and PaDiM to improve their efficiency in the continual learning setting.

Despite this growing interest, the vast majority of continual VAD methods have been developed and evaluated under the assumption of unconstrained computational resources. A first attempt to jointly address continual learning and edge deployment was made by \cite{barusco2025memory}, who evaluated STFPM and PaSTe in the continual edge setting.

However, these studies remain limited in scope. 
They focus on a small number of models and backbone architectures and do not provide a systematic analysis. In addition, it remains unclear which VAD architectures and backbone designs offer the best trade-off between efficiency, adaptability, and detection performance when both constraints are imposed simultaneously. Our work addresses this gap by providing the first comprehensive benchmark for continual VAD on the edge, covering seven VAD models across three lightweight backbone architectures, and by proposing new efficient solutions tailored to this challenging regime.

    
    

\section{Methodology}
\label{sec:methodology}

In this section, we describe the methods proposed to address the challenges of continual VAD on the edge. We begin by presenting Tiny-Dinomaly, a lightweight adaptation of the Dinomaly model designed to retain strong anomaly detection performance under strict computational constraints. We then introduce PaDiM MultiModal and PatchCore++, two targeted modifications of their respective baselines aimed at improving efficiency and suitability for continual learning on resource-constrained devices.

\subsection{Continual Learning for VAD} 
In real-world industrial deployments, the assumption of a stationary data distribution is rarely satisfied.
Manufacturing environments are inherently dynamic, with new products and process variations continuously altering the distribution of normal images.
Each change introduces a shift that needs to be addressed.

For example, a quality control system deployed on a production line initially inspects a single component type, e.g., metal nuts. Over the course of months, the system must be extended to cover cables, capsules, transistors, and further items as they enter production. A continual learning strategy that can incrementally absorb new tasks while preserving performance on previous ones is a practical necessity for scalable industrial inspection systems.

Following the formulation of previous works, we adopt a continual learning scenario in which each task $t \in \{1, \dots, T\}$ corresponds to a distinct object category (see Figure \ref{fig:cl_vad_scenario}). At task $t$, the model receives the the images of the t-th task, all of which depict normal (defect-free) samples, consistent with the unsupervised anomaly-detection paradigm.

After training on task $t$, the model is evaluated on all tasks seen so far. This evaluation protocol measures the capability of the model to generalize across a growing set of items while being constrained by fixed memory and computation budgets.
\\
\\
Most VAD methods developed for continual learning assume access to specific CL mechanisms, making them difficult to adapt to edge deployment scenarios. To enable fair and consistent evaluation across a broad range of classic VAD models, we require a CL strategy that is agnostic to model architecture and imposes minimal assumptions.
For this reason, we adopt the Replay approach to evaluate and compare VAD models in the CL setting. Replay is simple, efficient, and model-agnostic, making it well-suited to a fair comparison across heterogeneous VAD architectures.
In the Replay strategy, the model is trained jointly on data from the current task and a small subset of data retained from previous tasks. In our experiments, we analyze the effect of replay memory size by evaluating two configurations: 40 and 100 samples, corresponding to approximately 5 MB and 15 MB of memory, respectively. While a buffer of 40 samples represents the most realistic scenario for edge deployment, given its minimal memory footprint, we also assess a larger buffer size to study the impact of increased memory capacity on continual learning performance.
\\
\\
Replay can be seamlessly applied to any end-to-end VAD model. In this work, we evaluate four such models, STFPM, PaSTe, SimpleNet, and Dinomaly, each adapted for edge deployment by substituting the backbone with a lightweight alternative.
Specifically, we test each model across multiple edge-oriented backbones, namely MobileNetV2, MCUNet, and PhiNet, covering a range of memory and computational budgets typical of resource-constrained devices.

However, several widely adopted VAD methods, such as PatchCore, CFA, and PaDiM, do not follow an end-to-end training paradigm. Instead, they rely on auxiliary data structures, a memory bank in the case of PatchCore and CFA, and Gaussian distributions for PaDiM, which require dedicated strategies for continual learning. To handle these cases, we adopt the CL adaptations proposed in \cite{bugarin2024unveiling}, namely PatchCore-CL and PaDiM-CL. 

However, since none of the original models were designed with edge deployment in mind, we introduce novel edge-optimized variants that reduce computational and memory overhead while preserving the ability to retain knowledge of previously learned objects, as discussed in the next sections.

As an upper bound, we consider Joint Training (JT), where the model trains on all tasks simultaneously, serving as the ideal method that is unaffected by catastrophic forgetting. As a lower bound, Fine-Tuning (FT) trains exclusively on the current task with no forgetting mitigation, representing the worst-case baseline.

\subsection{VAD Models}

\begin{itemize}
    \item \textbf{PatchCore}~\cite{roth2022towards}: it builds a compact memory bank of representative normal patches and flags anomalies based on the distance of test patches to their nearest neighbors.
    \item \textbf{PaDiM}~\cite{defard2021padim}: it models each spatial feature location with a multivariate Gaussian and uses the Mahalanobis distance to detect deviations as anomalies.
    \item \textbf{PaDiM-Lite} \citep{stropeni2026efficient}: it is a lightweight variant of PaDiM that replaces the full covariance matrix with a diagonal approximation, retaining only per-dimension variances. This reduces memory per spatial position from $\mathcal{O}(d^2)$ to $\mathcal{O}(d)$ and eliminates the need for matrix inversion, making it significantly more efficient for edge deployment.
    \item \textbf{CFA}~\cite{9839549}: it creates a memory of normal patch embeddings and adapts them into coupled hyperspheres to amplify the separation between normal and abnormal feature representations.
    \item \textbf{STFPM}~\cite{wang2021studentteacherfeaturepyramidmatching}: it is based on two networks (teacher and student) with knowledge distillation, where student and teacher feature map deviations indicate anomalies.
    \item \textbf{PaSTe} \citep{barusco2025paste}: it is a resource-efficient variant  of STFPM that shares the first layers between teacher and student networks, reducing memory footprint and inference cost while preserving anomaly detection performance.
    \item \textbf{SimpleNet}~\cite{liu2023simplenet}: it also uses a feature adaptor like CFA, but also considers the generation of synthetic anomalies at feature-level to improve performance.
    \item \textbf{Dinomaly}~\cite{guo2025dinomaly}: it is a minimalistic reconstruction-based framework built on pure Transformer architectures, leveraging linear attention and a noisy bottleneck to enable a unified model for multi-class anomaly detection.
\end{itemize}

\subsection{Dinomaly}
Dinomaly follows the same encoder-decoder paradigm introduced by methods like RD4AD, but replaces the conventional CNN backbone with a fully transformer-based architecture. The encoder is a foundation model, specifically a Vision Transformer (ViT) pretrained via DINOv2 \cite{caron2021emerging}, serving as a powerful frozen feature extractor. 
The decoder is also a transformer, randomly initialized and trained from scratch on the target dataset, with the objective of reconstructing the intermediate feature representations produced by the foundation model.
At inference, the decoder reconstructs normal regions well but fails on anomalies, producing a discrepancy between encoder and decoder features that serves as the anomaly signal.
\\
The core challenge in reconstruction-based anomaly detection is identity mapping: the decoder learns to reconstruct everything too well, including anomalies, making detection impossible. Dinomaly addresses this through three components.
i) \textit{Noisy Bottleneck} applies Dropout to the MLP bottleneck, randomly discarding activations to prevent the decoder from perfectly memorizing and reproducing unseen patterns.
ii) \textit{Linear Attention} exploits a known weakness of linear attention mechanisms, their inability to focus on local regions, forcing the decoder to rely on diffuse, long-range information rather than directly copying local input tokens.
iii) \textit{Loose Reconstruction} relaxes the supervision strategy; instead of strict layer-to-layer feature matching, features from multiple encoder layers are grouped and reconstructed collectively, giving the decoder more freedom to diverge from the encoder when faced with unseen anomalous patterns.
\\
Compared to classic reconstruction-based approaches, Dinomaly is computationally efficient since the encoder is frozen and only the lightweight decoder is trained. Although originally introduced in a multi-class setting, the method is naturally compatible with continual learning, since replay can be applied without modifying its training objective.

\subsection{Tiny-Dinomaly}

While Dinomaly achieves state-of-the-art performance through its simple yet effective design choices, its size makes it infeasible for edge deployment. The foundation model ViT and the decoder together require 561.33 MB of storage (excluding the replay memory), making deployment on resource-constrained devices impractical.

To address this limitation, we propose \textit{Tiny-Dinomaly}, a lightweight variant that replaces the heavy DINOv2 ViT-L/14 backbone \cite{oquab2023dinov2} (~307M parameters) with DeiT-Tiny (~5M parameters). DeiT (Data-efficient Image Transformer) is a compact vision transformer that, unlike DINOv2, trains exclusively on ImageNet without requiring additional large-scale proprietary datasets, leveraging a knowledge distillation strategy to compensate for the reduced data budget \cite{touvron2021training}. Despite its smaller scale, DeiT-Tiny retains the same transformer backbone structure as ViT, preserving full compatibility with Dinomaly's reconstruction paradigm while drastically reducing the model footprint, making it suitable for deployment on resource-constrained devices.
The decoder follows the same design principle as Dinomaly: a transformer of comparable scale to the encoder, randomly initialized as in the original framework, and trained to reconstruct the encoder's intermediate features using the same Noisy Bottleneck, Linear Attention, and Loose Reconstruction strategies introduced in Dinomaly.
\\
\\
While Tiny-Dinomaly retains the Noisy Bottleneck and Linear Attention components from the original method, our architecture introduces an additional structural advantage. These components were designed to constrain the decoder's reconstruction power, preventing identity mapping. In Tiny-Dinomaly, a similar effect emerges naturally from the encoder itself: unlike large DINOv2 ViTs, which produce highly complete and semantically rich representations, a smaller encoder like DeiT-Tiny extracts inherently less exhaustive features. This makes it harder for the decoder to generalize to unseen anomalous patterns, providing a built-in inductive bias that complements, and partially replaces, the role of the explicit regularization components introduced in Dinomaly.

In other words, where Dinomaly artificially constrains feature completeness and reconstruction power through components like Noisy Bottleneck and Linear Attention, Tiny-Dinomaly achieves the same effect structurally. 
Furthermore, with fewer parameters to model the normal feature distribution, the decoder converges faster and more tightly to what "normal" looks like, rather than drifting toward a more general solution, improving sensitivity to subtle deviations.

Therefore, Tiny-Dinomaly may achieve a similar regularizing effect structurally, through the reduced representational capacity of the encoder itself. The resulting model is substantially more suitable for resource-constrained deployment and, as shown in Section~\ref{sec:results}, can also achieve stronger anomaly localization performance.

\subsection{PatchCore}
PatchCore extracts hierarchical representations using a pre-trained Convolutional Neural Network (CNN) backbone. Let $\phi$ denote the feature extraction network, for a given input image $x_i$, let $\phi_j(x_i)$ represent the feature map extracted from the $j$-th layer. 
For each spatial position $(h, w)$ of the feature map $\phi_{i,j} \in \mathbb{R}^{c^* \times h^* \times w^*}$, PatchCore aggregates features within a $p \times p$ local neighborhood via adaptive average pooling, yielding a single patch descriptor $\phi_{i,j}(h,w)$. Repeating this at every stride-$s$ position and across all training images $x_i \in X_N$ gives the memory bank:
$
\mathcal{M} = \bigcup_{x_i \in X_N} \mathcal{P}_{s,p}(\phi_j(x_i))
$.
\\
PatchCore finds a subset  $\mathcal{M}_C \subset \mathcal{M}$ that best preserves spatial coverage in feature space by applying the k-center algorithm.
\\
The anomaly score is driven by the single most anomalous patch:
\begin{equation}
    m^{test,*},\, m^* = \arg\max_{m^{test} \in \mathcal{P}(x^{test})} 
    \arg\min_{m \in \mathcal{M}} \|m^{test} - m\|_2, 
    \qquad
    s^* = \|m^{test,*} - m^*\|_2
\end{equation}
This raw score is reweighted to account for how typical $m^*$ itself is within $\mathcal{M}$:
\begin{equation}
    s = \left(1 - \frac{\exp\|m^{test,*} - m^*\|_2}
    {\sum_{m \in \mathcal{N}_b(m^*)} \exp\|m^{test,*} - m\|_2}\right) \cdot s^*
\end{equation}

where $\mathcal{N}_b(m^*)$ is the set of $b$ nearest neighbors of $m^*$ inside $\mathcal{M}$.
For anomaly segmentation, per-patch scores are placed back at their spatial positions and upsampled via bilinear interpolation, followed by Gaussian smoothing ($\sigma=4$).

\subsection{PatchCoreCL}
The core challenge is that standard CL techniques like replay do not directly apply to PatchCore, since its feature extractor weights are frozen; the only component that "learns" is the memory bank.
Instead of a single memory bank, PatchCoreCL maintains one separate memory bank per task, while keeping the total number of stored patch vectors fixed (either 10K or 30K). As new tasks arrive, the per-task budget shrinks as $\frac{\text{memory size}}{\text{number of tasks seen so far}}$.
When the task arrives:
\begin{enumerate}
    \item All existing memory banks are coreset-subsampled down to the new (smaller) per-task budget
    \item The new task's patches are also coreset-subsampled to the same budget
    \item The new memory bank is appended to the list
\end{enumerate}

This keeps total memory bounded while preserving the coverage quality of each task's representation.

At test time, the input patches are compared against every memory bank in the list. The one returning the lowest image-level anomaly score is selected as the matching task, and its memory bank is then used for pixel-level scoring. This also implicitly performs task identification without any explicit task label.

For complexity analysis, considering a vector operation as a unitary cost, it can be proved that the cost to update the memory banks when a new task appears has cost
$O\left( \frac{S(S+N)}{i} \right)$,
denoting by $S$ the total memory budget, by $N$ the number of raw patches from the new task, and by $i \in \{1, \ldots, T\}$ the index of the current task.

The complexity is attributed to two phases of the update process.
\textbf{Phase 1} updates the previous $i$ old memory banks with a total cost of $O\left( \frac{S^2}{i} \right)$ since it reduces $i$ memory banks from size $\frac{S}{i-1}$ to $\frac{S}{i}$ using the k-center algorithm.
\textbf{Phase 2} creates the new memory bank at cost $O\left( \frac{NS}{i} \right)$.
Therefore, the total update cost at task $i$ by summing phase 1 and phase 2 is $O\left( \frac{S(S+N)}{i} \right)$.

\subsection{PatchCoreCL++}
While PatchCoreCL achieves very low forgetting, two practical limitations remain. First, the cost of continuously pruning old memory banks at each new task is expensive. Second, inference time grows with the number of tasks, since test patches must be compared against all memory banks in the list.
In this variant, we propose a lighter training and a lighter inference procedure.

\textbf{Training} Therefore, we propose to replace the k-center recompression of Phase 1 with an efficient pruning strategy motivated by a classical result in combinatorial optimization. The k-center algorithm with farthest-first traversal \cite{gonzalez1985clustering} iteratively selects the point farthest from the current set of chosen centers, adding it to the coreset at each step. 
Authors of \cite{gonzalez1985clustering} proved that this greedy procedure enjoys a fundamental prefix property: for any $k$, the first $k$ points form a 2-approximation of the optimal k-center solution, meaning no other subset of size $k$ can cover the remaining points with a radius less than half that achieved by the greedy prefix. As a direct consequence, the ordering induced by farthest-first traversal is monotonically decreasing in representativeness: points inserted earlier cover broader, more spread-out regions of feature space, while points inserted later fill in progressively finer details in already well-covered regions.

Under this formulation, when task $i$ arrives and the per-task budget shrinks from 
$\frac{S}{i-1} \to \frac{S}{i}$, instead of rerunning the k-center 
algorithm on each old memory bank, which costs $O\left(\frac{S^2}{i}\right)$ in total, 
we simply truncate each list by removing the last $\frac{S}{i(i-1)}$ 
entries. By the prefix property of Gonzalez's algorithm, this corresponds to discarding 
precisely the least representative points first, preserving the coverage quality of the 
remaining prefix. This reduces the Phase 1 cost from $O\left(\frac{S^2}{i}\right)$ 
to $O(1)$, making the total update cost at task $i$ as $O\left(\frac{NS}{i}\right)$.

This allows PatchCoreCL++ to scale to larger memory banks more easily compared to the 
original PatchCoreCL. While PatchCoreCL grows quadratically in $S$, PatchCoreCL++ has complexity $O\left(\frac{NS}{i}\right)$, which is linear in $S$.

\textbf{Inference}
The original PatchCoreCL strategy scores a test image against every memory bank and selects the one returning the lowest anomaly score. Consequently, the inference time grows linearly with the number of tasks $K$: the cost of scoring a single test image is $O(K \cdot |M| \cdot d)$, where $|M|$ denotes the per-task memory bank size and $d$ represents the patch descriptor dimension. As the number of tasks grows, this linear factor renders PatchCoreCL increasingly impractical for deployment on edge devices where latency budgets are strict.

Instead, we propose a prototype-based task identification. For each task $j$, we define a prototype as the mean of the backbone feature maps extracted by the last feature extraction layer from its training images $X_N^{(j)}$:
\[
\mathbf{p}_j = \frac{1}{|X_N^{(j)}|} \sum_{x_i \in X_N^{(j)}} \phi(x_i)
\]
and represent a test image $x^{test}$ by its backbone feature map $\phi(x^{test})$. Task identification then reduces to a single nearest-prototype lookup:
\[
j^* = \arg\min_{j \in \{0,\ldots,K-1\}} \|\phi(x^{test}) - \mathbf{p}_j\|_2
\]

The assigned task $j^*$ determines which memory bank $\mathcal{M}[j^*]$ is used for anomaly scoring. Consequently, only one memory bank is queried at inference time rather than all $K$ banks. This approach reduces the inference cost from $O(K \cdot |M| \cdot d)$ to $O(K \cdot d)$ for the identification step, plus $O(|M| \cdot d)$ for the anomaly scoring. This represents a reduction by a factor of $K$ in the dominant term, an advantage that becomes increasingly pronounced as more tasks are accumulated.

It should be observed that the additional storage required for prototype-based identification is negligible. Maintaining $K$ prototypes requires only $K \times d$ scalar values, where $d$ denotes the dimension of the backbone's final feature layer. In contrast, the memory bank itself stores $K \cdot |M| \cdot d$ values. Given that $|M| \gg 1$, the prototype overhead is smaller by a factor of $|M|$, rendering the inference speedup effectively free in terms of memory requirements.

\subsection{Padim and Padim-Lite}

As for PatchCore, the input image is passed through a pretrained CNN producing activation maps at multiple layers. For each spatial position (i, j) in the grid defined by the largest activation map, PaDiM extracts the activation vectors from the first three layers and concatenates them into a single embedding vector.

For each position (i, j), the N training images contribute N embedding vectors. PaDiM collects them into a set:
$\mathcal{X}_{ij} = \{x_{ij}^k, \, k \in [[1, N]]\}$ and models this set as a multivariate Gaussian $\mathcal{N}(\mu_{ij}, \Sigma_{ij})$, where:
\begin{equation}
    \mu_{ij} = \frac{1}{N} \sum_{k=1}^{N} x_{ij}^k
\end{equation}

\begin{equation}
\Sigma_{ij} = \frac{1}{N-1} \sum_{k=1}^{N} (x_{ij}^k - \mu_{ij})(x_{ij}^k - \mu_{ij})^\top + \epsilon I.
\end{equation}

The regularization term $\varepsilon I$ (with $\varepsilon=0.01$) ensures $\Sigma_{ij}$ is full rank and invertible, which is required for the next step. The covariance matrix $\Sigma_{ij}$ captures not just the variance of each feature, but the correlations between features coming from different CNN layers.

Given a test image, PaDiM extracts patch embeddings the same way. For each position $(i, j)$, it computes the Mahalanobis distance between the test embedding $x_{ij}$ and the learned Gaussian:

\begin{equation}
M(x_{ij}) = \sqrt{(x_{ij} - \mu_{ij})^\top \, \Sigma_{ij}^{-1} \, (x_{ij} - \mu_{ij})}
\end{equation}

Unlike plain Euclidean distance, the Mahalanobis distance accounts for the shape and orientation of the learned distribution via $\Sigma^{-1}$. A patch that deviates along a direction of low variance gets penalized more heavily than one deviating along a high-variance direction.

\textbf{Padim-Lite}
\\
The full covariance matrix $\Sigma_{ij} \in \mathbb{R}^{d \times d}$ is the computational bottleneck of PaDiM. Storing it requires $4d^2$ bytes per spatial position, and inverting it costs $\mathcal{O}(d^3)$ at training time, while each Mahalanobis distance evaluation costs $\mathcal{O}(d^2)$ at inference.

PaDiM-Lite replaces the full covariance with a diagonal approximation, retaining only the per-dimension variances $
\sigma^2_{ij} = \operatorname{diag}(\Sigma_{ij}) = \left(\sigma^2_{ij,1}, \sigma^2_{ij,2}, \ldots, \sigma^2_{ij,d}\right)
$
where each component is estimated from the $N$ training embeddings at position $(i,j)$:
\begin{equation}
\sigma^2_{ij,k} = \frac{1}{N-1}\sum_{n=1}^{N}\left(x^{(n)}_{ij,k} - \mu_{ij,k}\right)^2 + \varepsilon
\end{equation}
with $\varepsilon = 0.01$ added for numerical stability (ensuring the denominator never vanishes). By assuming independence across feature dimensions, the Mahalanobis distance reduces to a simple normalised sum of squared residuals:
\begin{equation}
M(x_{ij}) = \sqrt{\sum_{k=1}^{d} \frac{\left(x_{ij,k} - \mu_{ij,k}\right)^2}{\sigma^2_{ij,k}}}
\end{equation}
This is computed entirely through element-wise operations, with no matrix inversion and no matrix-vector products. The resulting complexity is $O(d \cdot N)$ at training and $O(d)$ per patch at inference, down from $O(d^2 \cdot N + d^3)$ and $O(d^2)$ respectively. Memory per spatial position shrinks from $4d^2$ bytes (the full matrix) to $4d$ bytes (mean vector) plus $4d$ bytes (variance vector), a reduction of $O(d)$.

The diagonal approximation discards cross-dimensional correlations. This is a deliberate trade-off: those correlations improve distributional modeling but come at a cost that is prohibitive for edge hardware.

\subsection{Padim Variants for Continual Learning}
The first adaptation of Padim for CL, here referred to as PadimCL, is in \cite{bugarin2024unveiling}.
The core problem is straightforward: standard PaDiM stores one Gaussian $\mathcal{N}(\mu_{ij}, \Sigma_{ij})$ per patch position per task. If you have $T$ tasks, memory grows as $O(T \times W \times H \times d^2)$, which violates any reasonable memory budget for continual learning. 

The proposed fix is to maintain a single Gaussian per patch position at all times, and update it incrementally as new tasks arrive, so memory stays $O(W \times H \times d^2)$ regardless of how many tasks have been seen.
Original method performs it as a simple arithmetic mean of the per-task estimates, not the statistically correct update. In mathematical terms, the code computes $ \mu_{(T)} = \frac{1}{T} \sum_{t=1}^{T} \mu_{ij}^{t} $ and $ \Sigma^{-1}_{(T)} = \frac{1}{T} \sum_{t=1}^{T} \Sigma^{-1,t}_{ij}  $.

However, the original method assigns a uniform weight of $1/T$ to each task when computing the mean and covariance, regardless of the number of samples it contains. This can introduce significant bias in both estimates when tasks have unequal dataset sizes and the resulting covariance does not correspond to the actual variability of the aggregated data.

In addition, the correct formula for the unified covariance requires operating on the covariance matrices , applying the correction terms $\delta^t {\delta^t}^\top$ to account for the shift between means, and only performing the inversion at the final step. 

\textbf{PadimCL-UniModal}
Therefore, we fix the original PadimCL method.
Suppose after $T-1$ tasks, position $(i,j)$ is represented by $\mathcal{N}(\mu^{(T-1)}_{ij}, \Sigma^{(T-1)}_{ij})$, estimated from $N_{T-1}$ total samples. A new task $T$ arrives with $N_T$ new samples, contributing a local Gaussian $\mathcal{N}(\mu^{T}_{ij}, \Sigma^{T}_{ij})$.

The updated mean is a weighted average:
\begin{equation}
\mu^{(T)}_{ij} = \frac{N_{T-1}\mu^{(T-1)}_{ij} + N_T\mu^{T}_{ij}}{N_{T-1} + N_T}
\end{equation}

The updated covariance uses the parallel axis theorem, which accounts both for within-task spread and the shift between task means:
\begin{equation}
\Sigma^{(T)}_{ij} = \frac{N_{T-1} \left( \Sigma^{(T-1)}_{ij} + \delta^{(T-1)}\delta^{(T-1)\top} \right) + N_T \left( \Sigma^{T}_{ij} + \delta^{T}\delta^{T\top} \right)}{N_{T-1} + N_T}
\end{equation}

where $\delta^{(T-1)} = \mu^{(T-1)}_{ij} - \mu^{(T)}_{ij}$ and $\delta^{T} = \mu^{T}_{ij} - \mu^{(T)}_{ij}$ are the deviations of each component mean from the new combined mean. These correction terms ensure that the merged covariance correctly reflects the spread of all samples seen so far, not just the spread within each task individually.

After the update, the old Gaussian is discarded. Only the new $(\mu^{(T)}_{ij}, \Sigma^{(T)}_{ij})$ and the total sample count $N_{T-1} + N_T$ are kept.

The resulting single Gaussian is the exact maximum likelihood estimate you would have obtained had you trained on all $N_{T-1} + N_T$ samples jointly — there is no approximation.

\textbf{PadimCL-MultiModal}
PaDiM-CL MultiModal forgoes the incremental fusion of PaDiM-CL UniModal and instead retains a separate Gaussian $\mathcal{N}(\mu^t_{ij}, \Sigma^t_{ij})$ per patch position for each task $t$. At inference time, the anomaly score is computed against every stored Gaussian, and the minimum score is taken as the final prediction:

\begin{equation}
M(x_{ij}) = \min_{t \in \{1,\ldots,T\}} \sqrt{(x_{ij} - \mu^t_{ij})^\top {\Sigma^t_{ij}}^{-1} (x_{ij} - \mu^t_{ij})}
\end{equation}

This is equivalent to selecting the task whose learned distribution best explains the test patch --- a natural fit for a continual learning setting where different tasks may correspond to genuinely distinct normal-appearance modes.

The cost of this expressiveness is that memory grows as $\mathcal{O}(T \times W \times H \times d^2)$, i.e., linearly with the number of tasks and without bound. This makes the method unsuitable for long task sequences on memory-constrained hardware, but it serves a precise role in this work: since each task's distribution is preserved exactly and never merged or approximated, PaDiM-CL MultiModal represents the performance ceiling for any PaDiM-based continual learning method. Any gap between it and PaDiM-CL UniModal quantifies the cost of enforcing a fixed memory budget through Gaussian fusion.

We propose lite variants of both continual learning formulations introduced above. \textbf{PaDiM-CL-Lite UniModal} maintains a single mean vector $\mu_{ij}$ and variance vector $\sigma^2_{ij}$ per patch position, updated incrementally across tasks using the same weighted fusion rule as PaDiM-CL UniModal, but operating on diagonal statistics only. \textbf{PaDiM-CL-Lite MultiModal} retains a separate mean and variance vector per task and position, selecting the best-matching task at inference by minimising the diagonal Mahalanobis score across all stored distributions.

The memory implications of the diagonal approximation are particularly significant in the multimodal setting. PaDiM-CL MultiModal must store, for each patch position, $T$ full covariance matrices of size $d \times d$, giving a total of $\mathcal{O}(T \times d^2)$ parameters per position. PaDiM-CL-Lite MultiModal reduces this to $\mathcal{O}(T \times d)$, since only the variance vector needs to be retained per task. Because $d$ is typically in the hundreds, the quadratic dependence on $d$ in the full-covariance case makes PaDiM-CL MultiModal substantially more expensive than its lite counterpart --- to the point where, for large enough $d$, PaDiM-CL UniModal (which stores a single $d \times d$ matrix) can exceed the memory cost of PaDiM-CL-Lite MultiModal (which stores $T$ vectors of length $d$).

This inversion of the usual memory ordering between unimodal and multimodal variants means that the optimal choice is not obvious \emph{a priori} and depends on the specific values of $T$ and $d$ in a given deployment. We therefore include an empirical analysis to characterise the performance--memory trade-off across all four variants.

\begin{figure}[!thbp]
    \centering
    \includegraphics[width=1\textwidth]{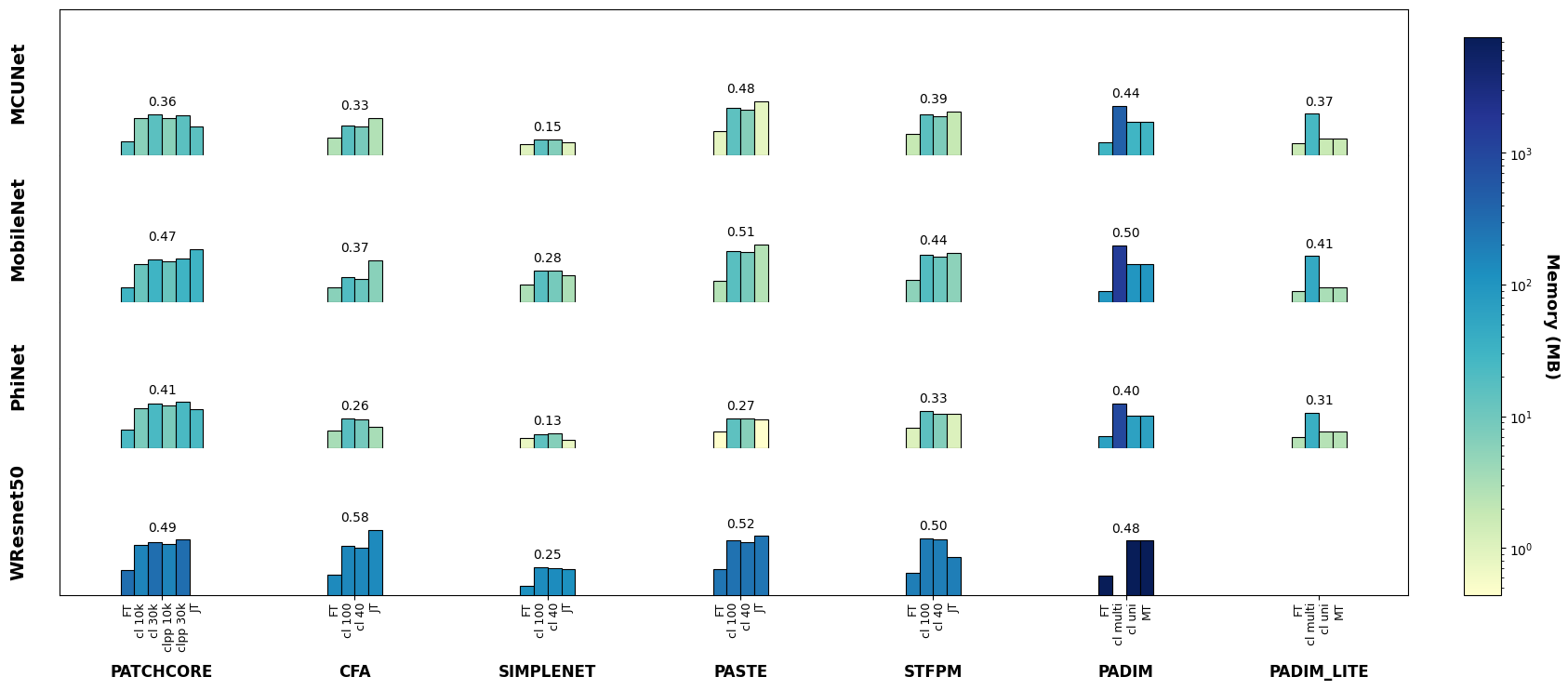}
    \caption{Performance (Pixel F1) and memory usage across different backbones and anomaly detection algorithms. Reported values correspond to the best performance achieved among the evaluated strategies. The color scale encodes memory consumption (in MB) on a logarithmic scale. FT denotes fine-tuning, while JT refers to joint training. “cl 10k” and “cl 30k” indicate memory bank sizes, whereas “cl 40” and “cl 100” denote replay memory sizes. “cl multi” and “cl uni” refer to PaDiM-based variants.}
    \label{fig:backbone_comparison}
\end{figure}

\section{Experimental Settings}
\label{sec:exp_settings}

\subsection{Evaluation Metrics}

\textbf{Visual Anomaly Detection Metrics}
We evaluate anomaly detection performance at both the image and pixel levels using the F1 score, a robust metric in the anomaly detection setting for highly imbalanced datasets.

\textbf{Continual Learning Metrics}
All VAD metrics are evaluated within a class-incremental continual learning setting. 
After training on the full sequence of tasks, we report the average final performance across all tasks. 

\textbf{Efficiency Metrics}
To assess suitability for edge deployment, we report two efficiency indicators. 
The \emph{memory footprint} is defined as the total storage required by model parameters and by any task-retention mechanism, including replay buffers, memory banks, or distribution statistics when applicable. 
The \emph{inference cost} is measured in GFLOPs per image and computed as the number of floating-point operations required for a forward pass.
\subsection{Datasets}

\textbf{MVTec AD.}
MVTec AD is the de facto standard benchmark for industrial anomaly detection~\cite{bergmann2019mvtec}. 
It comprises 5,354 high-resolution color images across 15 categories, including 10 object classes (e.g., bottle, cable, capsule, metal nut) and 5 texture classes (e.g., carpet, grid, leather). 

\textbf{VisA.}
VisA was introduced as a larger and more challenging complement to MVTec AD~\cite{zou2022spot}. 
It contains 10,821 high-resolution images across 12 object categories grouped into three domains.

\subsection{Experimental Protocol}

Models are trained sequentially over all categories and evaluated after each task on all categories observed so far.  All experiments were conducted using the MoViAD library \cite{barusco2025moviad}.
We consider replay buffer sizes of 40 and 100 images, reflecting realistic memory budgets in edge continual learning scenarios. 
For memory-bank methods such as PatchCore, we evaluate bank sizes of 10,000 (10k) and 30,000 (30k) entries.
We evaluate multiple lightweight backbones, including MobileNetV2 \cite{sandler2018mobilenetv2}, MCUNet \cite{lin2020mcunet}, and PhiNet \cite{phinet}, as well as WideResNet50 \cite{zagoruyko2016wide} as a high-capacity reference. 
For transformer-based models, Tiny-Dinomaly employs DeiT-Tiny.

We compare all methods under three training strategies: 
(i) Fine-Tuning (FT), where the model is trained only on the current task; 
(ii) Replay, where a subset of past samples is retained; 
and (iii) Joint Training (JT), where all data are available simultaneously and which serves as an upper performance bound.

\begin{figure*}[htbp]
    \centering
    \begin{subfigure}[t]{0.33\textwidth}
        \centering
        \includegraphics[width=\linewidth]{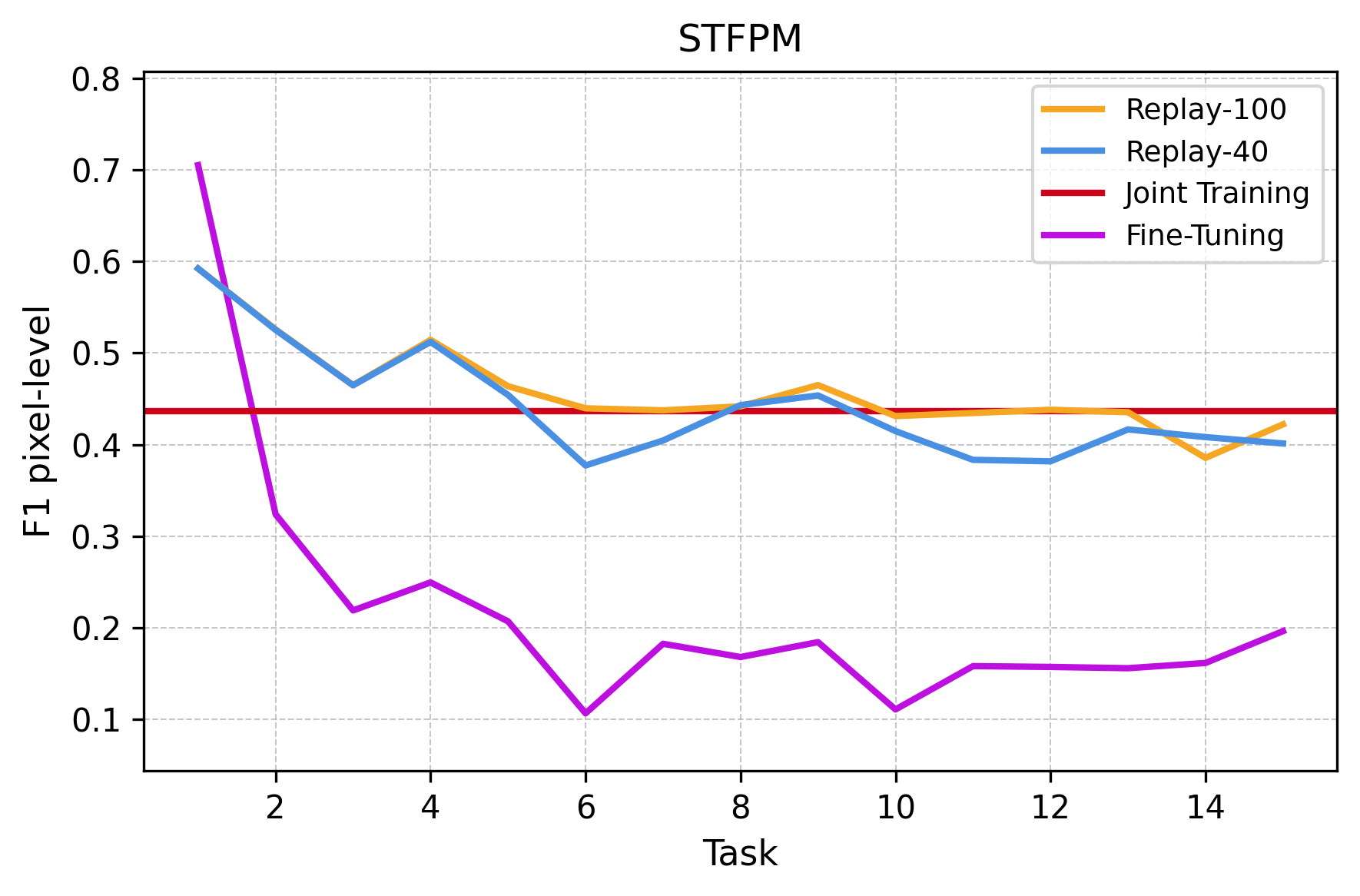}
        \caption{STFPM}
    \end{subfigure}\hfill
    \begin{subfigure}[t]{0.33\textwidth}
        \centering
        \includegraphics[width=\linewidth]{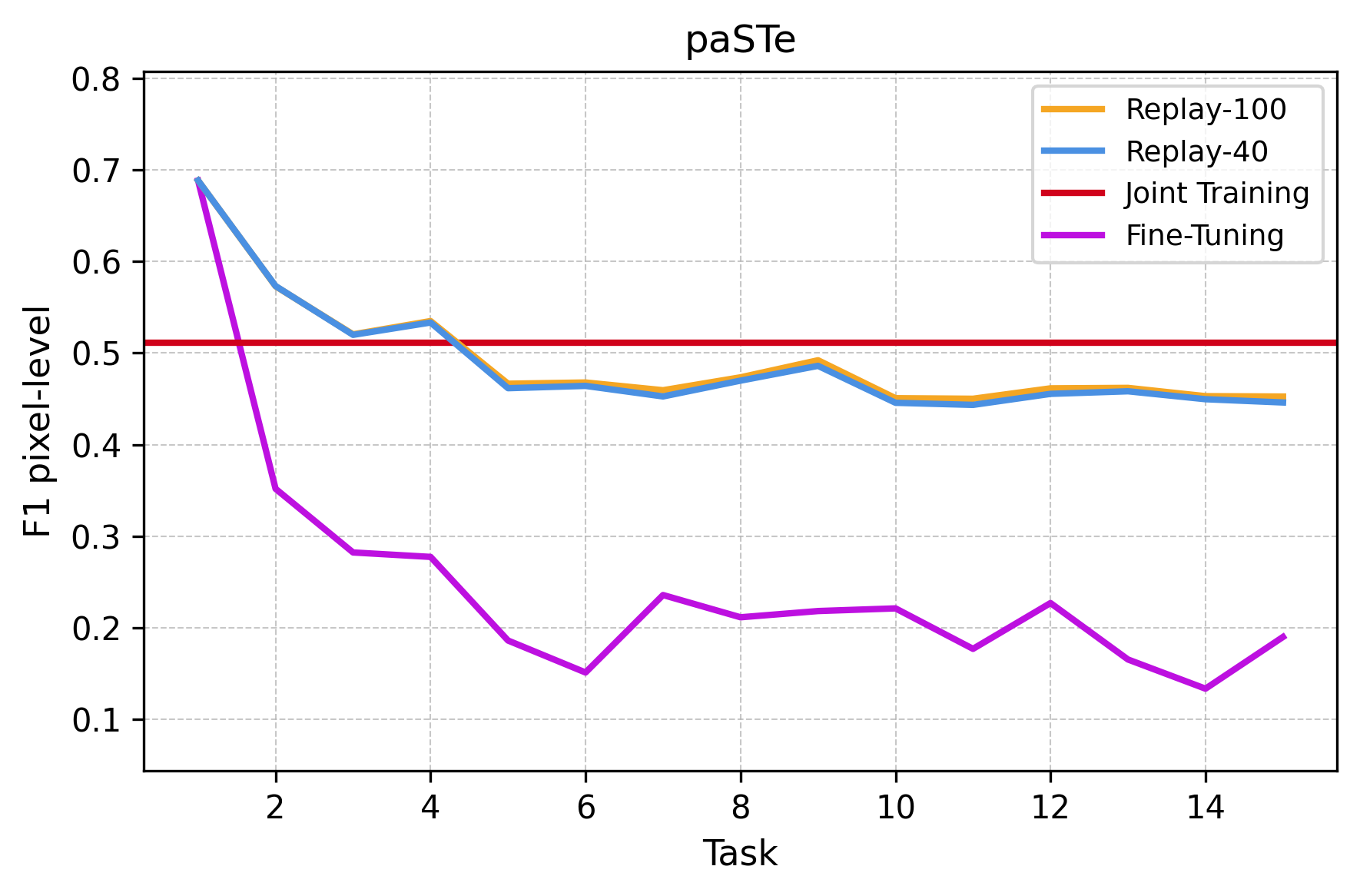}
        \caption{paSTe}
    \end{subfigure}\hfill
    \begin{subfigure}[t]{0.33\textwidth}
        \centering
        \includegraphics[width=\linewidth]{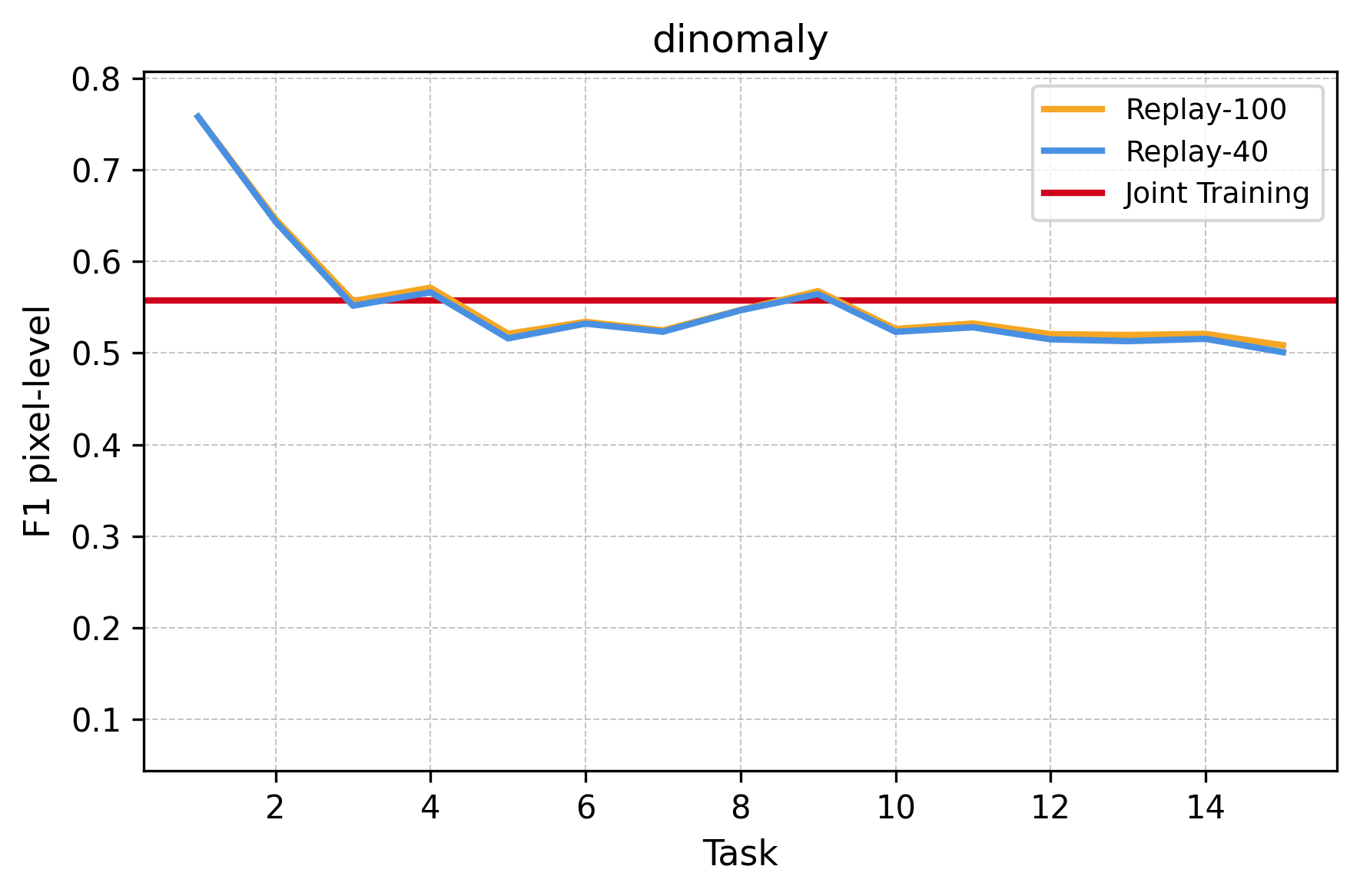}
        \caption{Tiny-Dinomaly}
    \end{subfigure}
    \vspace{0.5em}
    \begin{subfigure}[t]{0.33\textwidth}
        \centering
        \includegraphics[width=\linewidth]{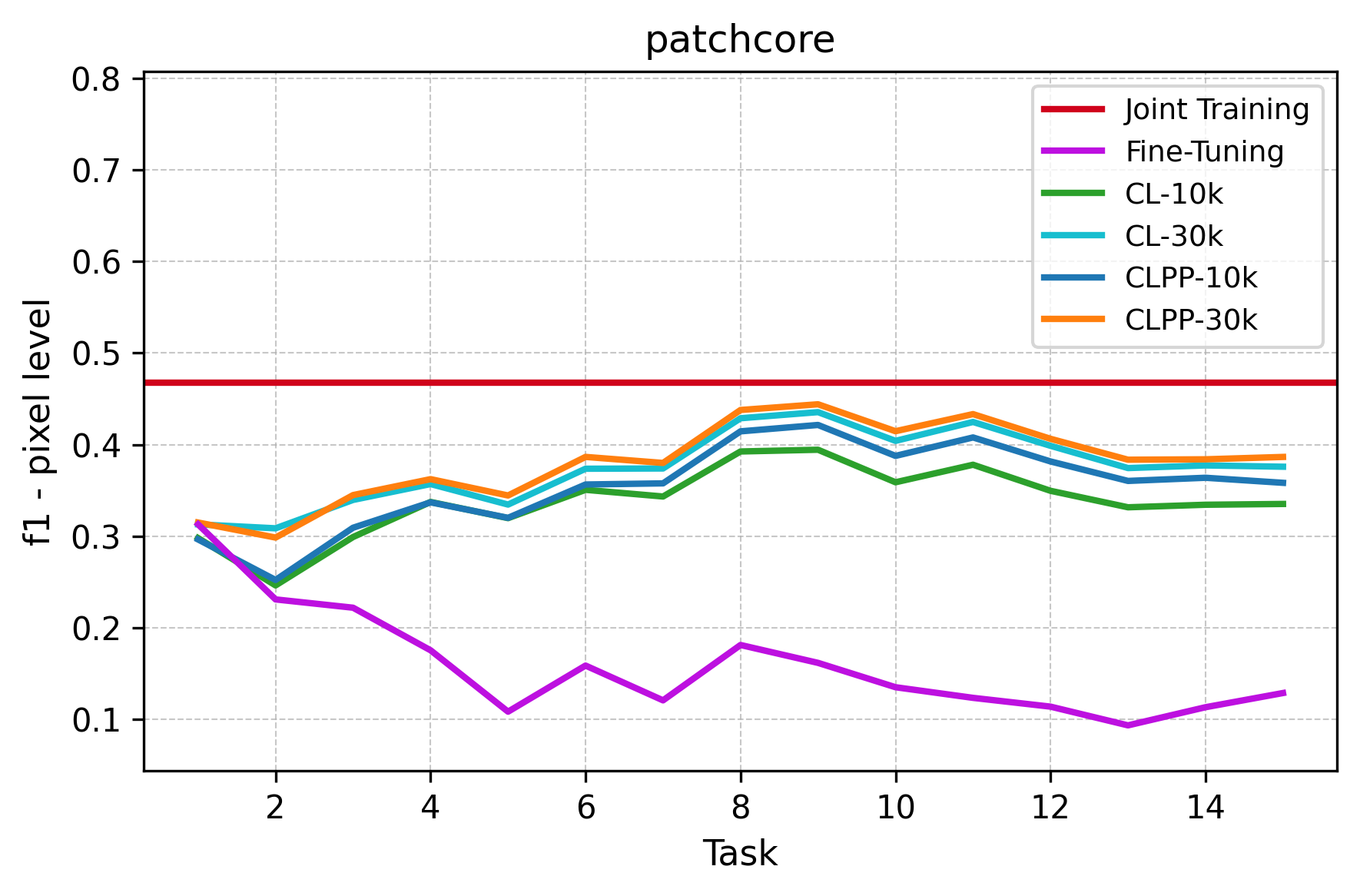}
        \caption{PatchCoreCL}
    \end{subfigure}\hfill
    \begin{subfigure}[t]{0.33\textwidth}
        \centering
        \includegraphics[width=\linewidth]{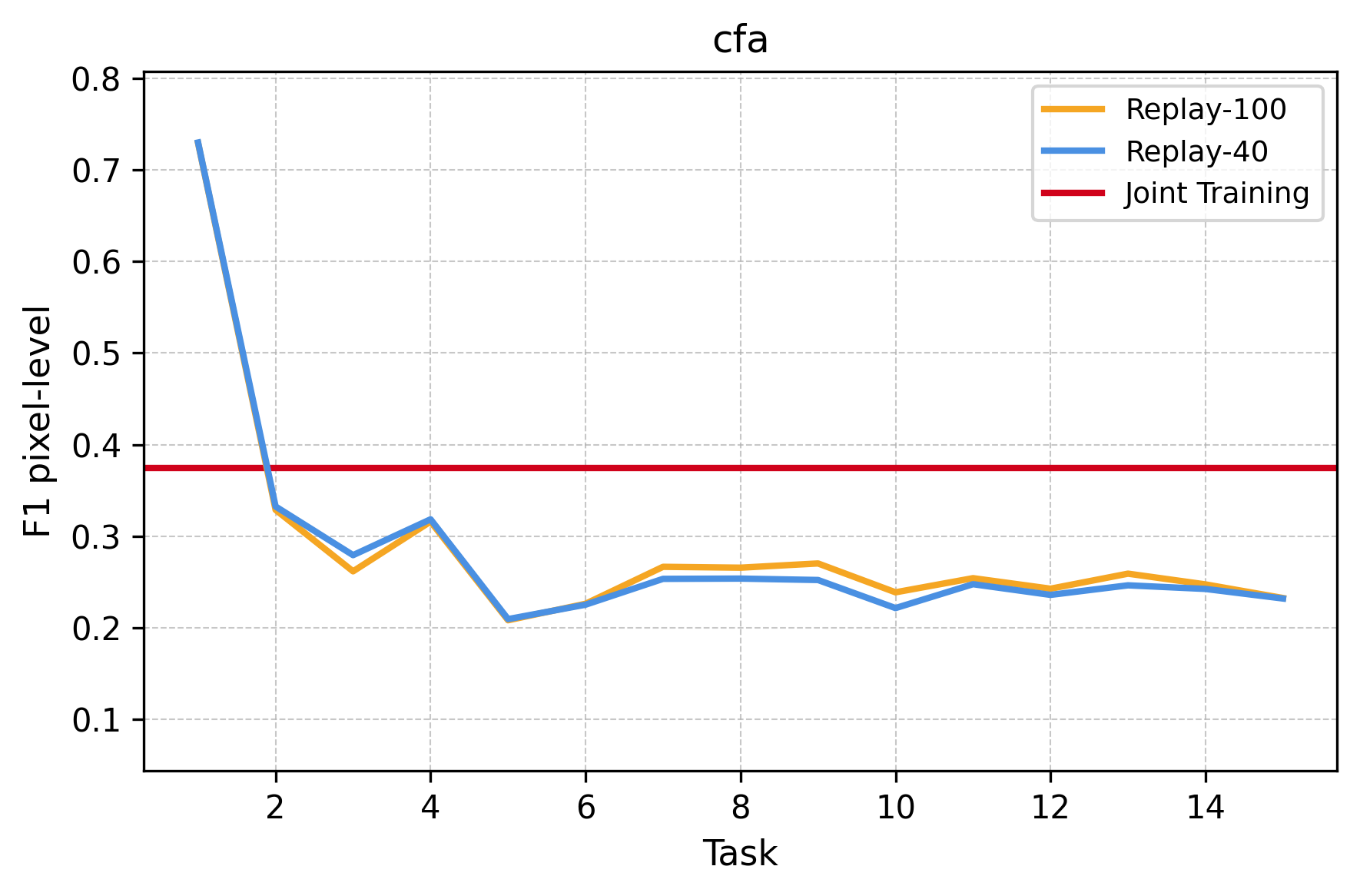}
        \caption{CFA}
    \end{subfigure}\hfill
    \begin{subfigure}[t]{0.33\textwidth}
        \centering
        \includegraphics[width=\linewidth]{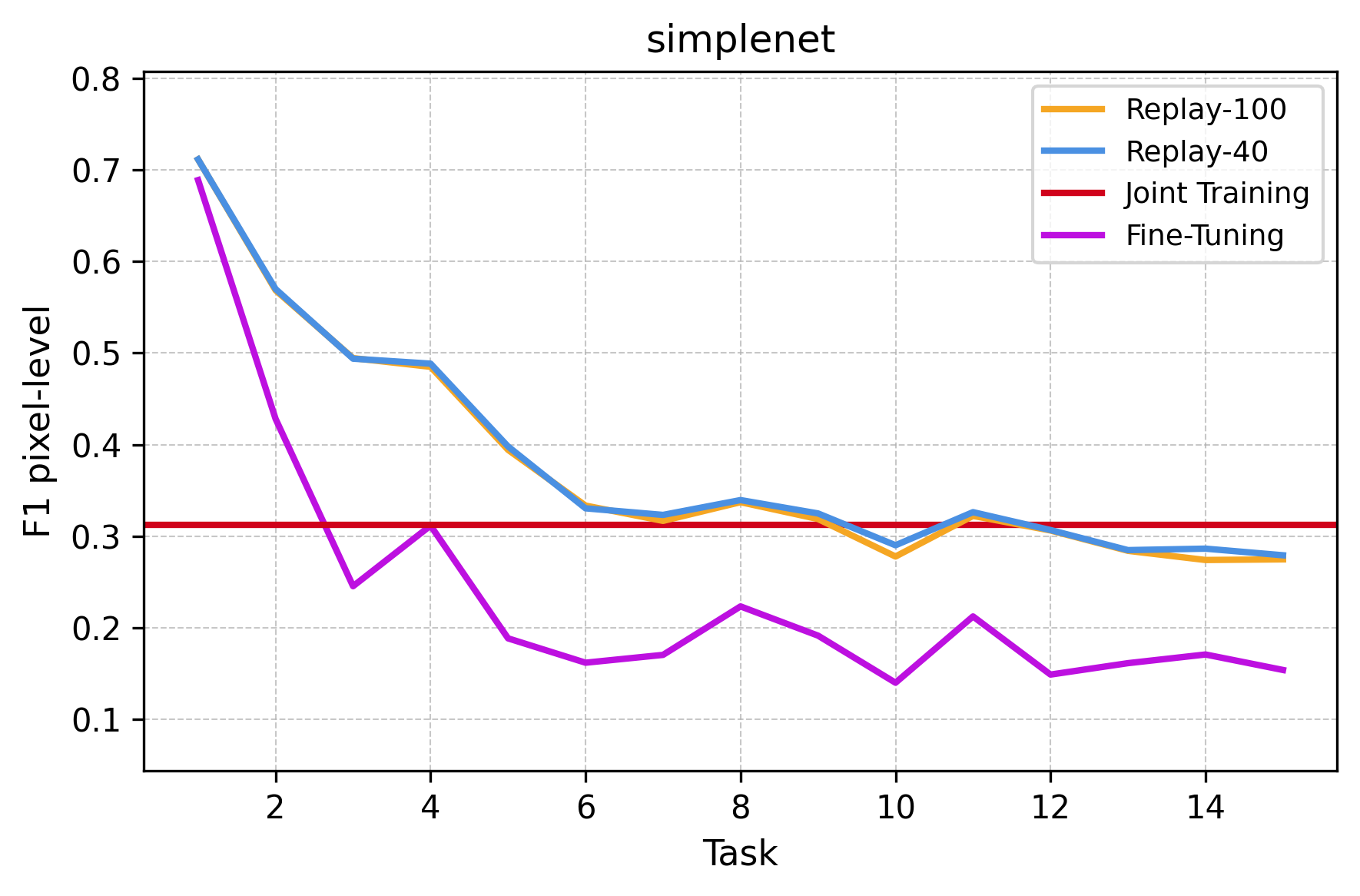}
        \caption{SimpleNet}
    \end{subfigure}
    \vspace{0.5em}
    \makebox[\textwidth][c]{%
        \begin{minipage}{0.63\textwidth}
            \centering
            \begin{subfigure}[t]{0.47\textwidth}
                \centering
                \includegraphics[width=\linewidth]{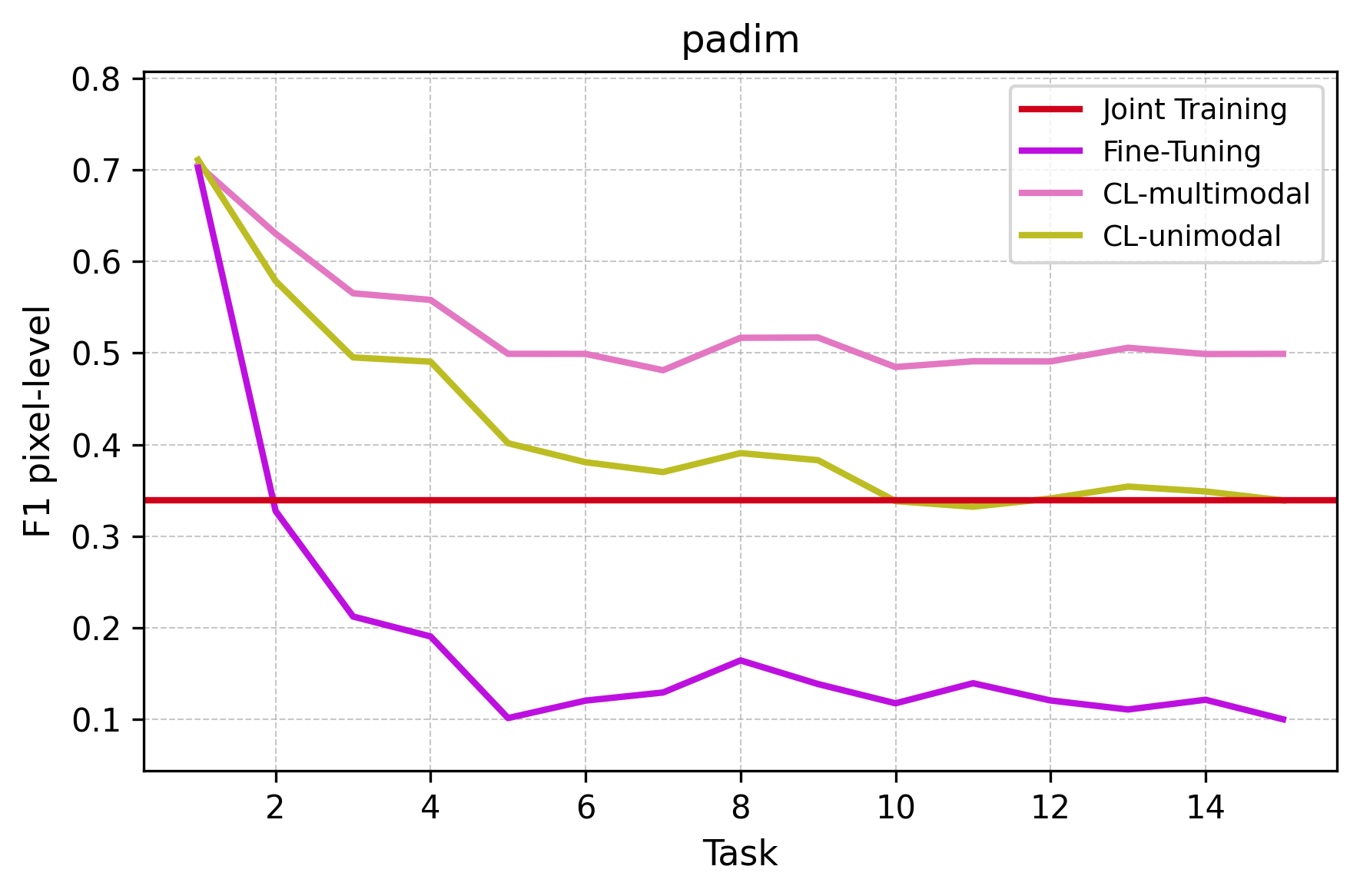}
                \caption{PaDiM}
            \end{subfigure}\hfill
            \begin{subfigure}[t]{0.47\textwidth}
                \centering
                \includegraphics[width=\linewidth]{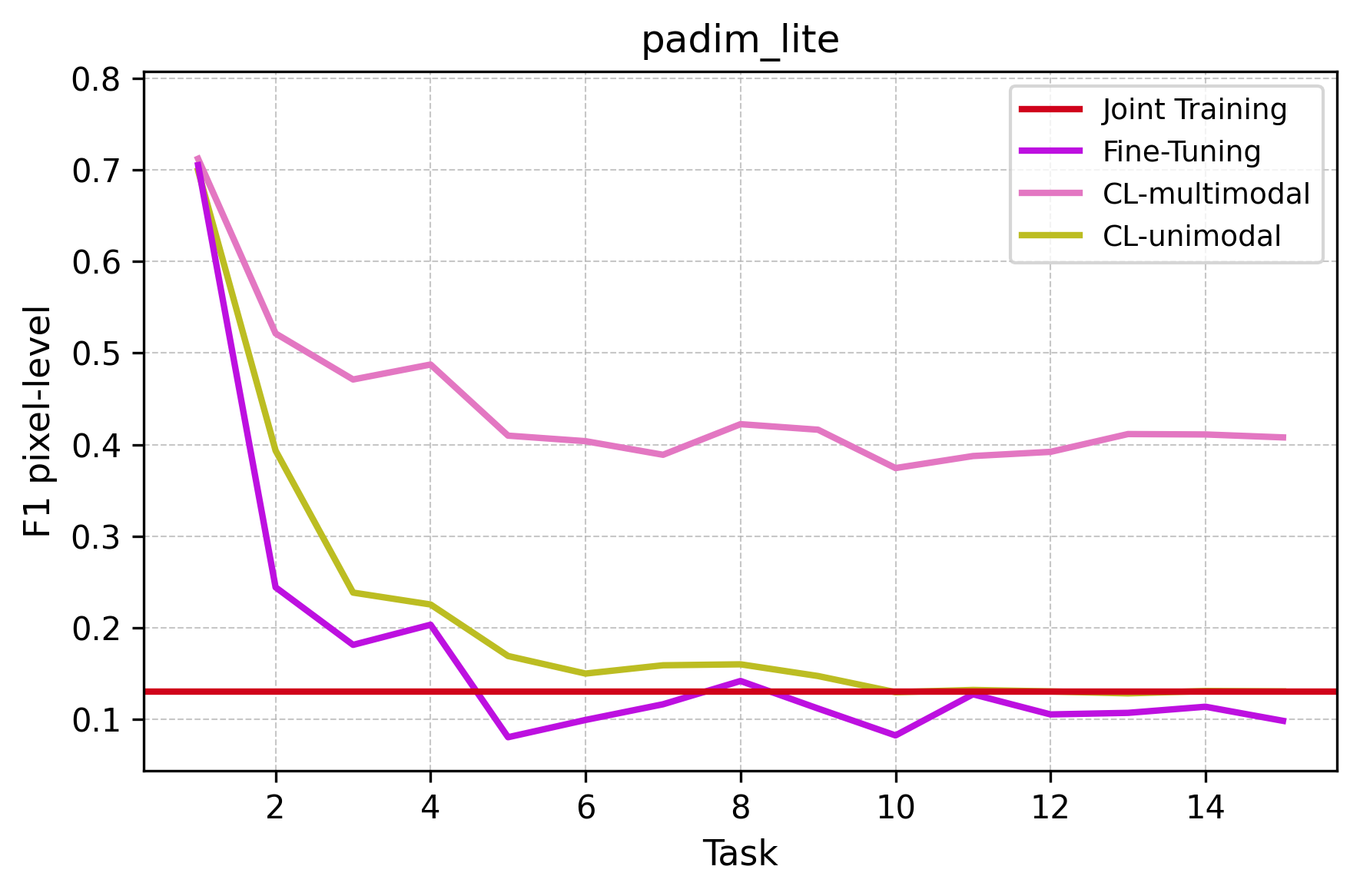}
                \caption{PaDiM Lite}
            \end{subfigure}
        \end{minipage}
    }
    \caption{Comparison of the considered methods across tasks in the CL setting using Mobilenet as backbone and MVTec as dataset.}
    \label{fig:cl_mobilenet_mvtec}
\end{figure*}

\section{Results}
\label{sec:results}

\begin{table}[th]
  \centering
  \small
  \caption{Comparison between the original PadimCL \cite{bugarin2024unveiling} and our version (here using WideResNet as backbone). Given the significant improvement in the results of our version we refer to our version as PadimCL in the rest of the paper.}
  \label{tab:padimcl_comparison}
  \begin{tabular}{lcccc}
    \toprule
    \textbf{Model} & \textbf{Image ROC} & \textbf{Image F1} & \textbf{Pixel ROC} & \textbf{Pixel F1} \\
    \midrule
    PaDiM CL (original)     & 0.56   & 0.80   & 0.73   & 0.17   \\
    PaDiM CL (ours) & 0.90 & 0.94 & 0.96 & 0.48 \\
    \bottomrule
  \end{tabular}
\end{table}

\subsection{Benchmark}

\begin{table}[!ht]
\caption{\textbf{Benchmark results on MVTec AD} across all model--backbone combinations in the continual learning setting. Results are reported as: \textit{Image F1 | Pixel F1 | GFLOPs | Memory (MB)}. cl 40 and cl 100 denote replay memory sizes of 40 and 100, respectively. cl multi refers to PaDiM-CL MultiModal, while cl uni denotes PaDiM-CL UniModal. CL 10K and CL 30K correspond to PatchCoreCL configurations with total memory bank sizes of 10,000 and 30,000 patches, respectively. Finally, CL++ indicates our proposed PatchCoreCL++ method.
}
\label{tab:anomaly_results}
\centering
\footnotesize
\setlength{\tabcolsep}{3pt}
\begin{tabular}{ll ccc}
\toprule
\textbf{Model} & \textbf{Strategy} & \textbf{MCUNet} & \textbf{MobileNet} & \textbf{PhiNet} \\
\midrule
\multirow{2}{*}{CFA} & cl 100 & 0.87 | 0.26 | 3.03 | 17.01 & 0.85 | 0.22 | 5.73 | 20.46 & 0.85 | 0.26 | 4.60 | 17.58 \\
 & cl 40 & 0.86 | 0.26 | 3.03 | 8.39 & 0.85 | 0.21 | 5.73 | 11.84 & 0.85 | 0.26 | 4.60 | 8.97 \\
\cmidrule(lr){1-5}
\multirow{2}{*}{PaDiM} & cl multi & 0.93 | 0.44 | 2.24 | 447.82 & 0.94 | 0.50 | 11.96 | 1469.91 & 0.91 | 0.40 | 6.52 | 959.62 \\
 & cl uni & 0.88 | 0.30  0.36 | 29.86 & 0.89 | 0.34 | 1.22 | 98.00 & 0.86 | 0.29 | 0.81 | 63.98 \\
\cmidrule(lr){1-5}
\multirow{2}{*}{PaDiM Lite} & cl multi & 0.91 | 0.37 | 0.34 | 25.59 & 0.90 | 0.41 | 0.57 | 46.67 & 0.88 | 0.31 | 0.52 | 37.63 \\
 & cl uni & 0.85 | 0.15 | 0.23 | 1.71 & 0.84 | 0.13 | 0.46 | 3.11 & 0.84 | 0.15 | 0.41 | 2.51 \\
\cmidrule(lr){1-5}
\multirow{4}{*}{PatchCore} & cl 10k & 0.95 | 0.33 | 10.52 | 6.06 & 0.93 | 0.34 | 17.77 | 12.11 & 0.94 | 0.35 | 14.72 | 8.14 \\
 & cl 30k & 0.96 | 0.36 | 27.84 | 16.45 & 0.95 | 0.38 | 49.16 | 31.04 & 0.96 | 0.40 | 40.08 | 23.41 \\
 & clpp 10k & 0.95 | 0.33 | 1.14 | 6.06 & 0.95 | 0.36 | 2.06 | 12.11 & 0.96 | 0.38 | 1.77 | 8.14 \\
 & clpp 30k & 0.96 | 0.36 | 2.29 | 16.45 & 0.96 | 0.39 | 4.15 | 31.04 & 0.96 | 0.41 | 3.46 | 23.41 \\
\cmidrule(lr){1-5}
\multirow{2}{*}{STFPM} & cl 100 & 0.90 | 0.37 | 0.45 | 16.19 & 0.91 | 0.42 | 0.90 | 19.82 & 0.87 | 0.33 | 0.81 | 15.43 \\
 & cl 40~~ & 0.89 | 0.34 | 0.45 | 7.58 & 0.90 | 0.40 | 0.90 | 11.21 & 0.87 | 0.31 | 0.81 | 6.82 \\
\cmidrule(lr){1-5}
\multirow{2}{*}{SimpleNet} & cl 100 & 0.84 | 0.14 | 0.53 | 15.36 & 0.87 | 0.27 | 1.20 | 17.41 & 0.84 | 0.13 | 0.93 | 15.14 \\
 & cl 40 & 0.84 | 0.15 | 0.53 | 6.74 & 0.87 | 0.28 | 1.20 | 8.80 & 0.84 | 0.13 | 0.93 | 6.52 \\
\cmidrule(lr){1-5}
\multirow{2}{*}{paSTe} & cl 100 & 0.91 | 0.42 | 0.32 | 15.22 & 0.92 | 0.45 | 0.68 | 16.94 & 0.86 | 0.27 | 0.56 | 14.80 \\
 & cl 40 & 0.90 | 0.41 | 0.32 | 6.60 & 0.92 | 0.45 | 0.68 | 8.32 & 0.86 | 0.26 | 0.56 | 6.18 \\
\bottomrule
\end{tabular}
\end{table}

\begin{table}[ht]
\centering
\caption{\textbf{Benchmark results on MVTec AD} for the Dinomaly model in the continual learning setting. Results are reported as: \textit{Image F1 | Pixel F1 | GFLOPs | Memory (MB)}.}
\label{tab:dinomaly_results}
\small
\begin{tabular}{ll cc}
\toprule
\textbf{AD Model} & \textbf{Strategy} & \textbf{DeiT Tiny} & \textbf{ViT Base} \\
\midrule
\multirow{3}{*}{Dinomaly} & cl 100 & 0.88 | 0.45 | 3.39 | 50.87 & 0.89 | 0.40 | 68.56 | 575.66 \\
 & cl 40  & 0.88 | 0.44 | 3.39 | 42.25 & 0.90 | 0.42 | 68.56 | 567.04 \\
\bottomrule
\end{tabular}
\label{tab:anomaly_results_dino}
\end{table}

\begin{table}[!htbp]
\centering
\caption{\textbf{Benchmark results on ViSA} across all model--backbone combinations in the continual learning setting. Results are reported as: \textit{Image F1 | Pixel F1 | GFLOPs | Memory (MB)}.}
\label{tab:ad_models_updated}
\footnotesize
\setlength{\tabcolsep}{3pt}
\begin{tabular}{@{} ll ccc @{}}
\toprule
\textbf{AD Model} & \textbf{Strategy} & \textbf{MCUNet} & \textbf{MobileNet} & \textbf{PhiNet} \\
\midrule

\multirow{2}{*}{CFA} 
 & cl 100 & 0.76 | 0.16 | 3.03 | 17.01 & 0.75 | 0.20 | 5.73 | 20.46 & 0.73 | 0.13 | 4.60 | 17.58 \\
 & cl 40  & 0.76 | 0.12 | 3.03 | \phantom{0}8.39  & 0.75 | 0.17 | 5.73 | 11.84 & 0.72 | 0.10 | 4.60 | \phantom{0}8.97 \\
\addlinespace

\multirow{2}{*}{PaDiM} 
 & cl multi & 0.83 | 0.24 | 2.24 | 447.82 & 0.84 | 0.29 | 11.96 | 1469.91 & 0.82 | 0.24 | 6.52 | 959.62 \\
 & cl uni   & 0.77 | 0.17 | 0.36 | 29.86  & 0.77 | 0.22 | \phantom{0}1.22 | \phantom{0}98.00  & 0.76 | 0.13 | 0.81 | \phantom{0}63.98 \\
\addlinespace

\multirow{2}{*}{PaDiM Lite} 
 & cl multi & 0.79 | 0.18 | 0.34 | 25.59 & 0.79 | 0.21 | 0.57 | 46.67 & 0.77 | 0.14 | 0.52 | 37.63 \\
 & cl uni   & 0.74 | 0.07 | 0.23 | \phantom{0}1.71  & 0.73 | 0.04 | 0.46 | \phantom{0}3.11  & 0.74 | 0.04 | 0.41 | \phantom{0}2.51 \\
\addlinespace

\multirow{2}{*}{PatchCore} 
 & clpp 10k & 0.85 | 0.05 | 1.14 | \phantom{0}6.06 & 0.86 | 0.06 | 2.06 | 12.11 & 0.86 | 0.06 | 1.77 | \phantom{0}8.14 \\
 & clpp 30k & 0.88 | 0.07 | 2.29 | 16.45 & 0.88 | 0.07 | 4.15 | 31.04 & 0.89 | 0.07 | 3.46 | 23.41 \\
\addlinespace

\multirow{2}{*}{STFPM} 
 & cl 100 & 0.76 | 0.17 | 0.45 | 16.19 & 0.79 | 0.20 | 0.90 | 19.82 & 0.74 | 0.12 | 0.81 | 15.43 \\
 & cl 40  & 0.77 | 0.18 | 0.45 | \phantom{0}7.58  & 0.80 | 0.24 | 0.90 | 11.21 & 0.73 | 0.02 | 0.81 | \phantom{0}6.82 \\
\addlinespace

\multirow{2}{*}{SimpleNet} 
 & cl 100 & 0.73 | 0.07 | 0.53 | 15.36 & 0.74 | 0.14 | 1.20 | 17.41 & 0.72 | 0.06 | 0.93 | 15.14 \\
 & cl 40  & 0.73 | 0.05 | 0.53 | \phantom{0}6.74  & 0.73 | 0.14 | 1.20 | \phantom{0}8.80  & 0.73 | 0.06 | 0.93 | \phantom{0}6.52 \\
\addlinespace

\multirow{2}{*}{paSTe} 
 & cl 100 & 0.77 | 0.24 | 0.32 | 15.22 & 0.78 | 0.28 | 0.68 | 16.94 & 0.73 | 0.02 | 0.56 | 14.80 \\
 & cl 40  & 0.78 | 0.20 | 0.32 | \phantom{0}6.60  & 0.79 | 0.25 | 0.68 | \phantom{0}8.32  & 0.73 | 0.04 | 0.56 | \phantom{0}6.18 \\
\bottomrule
\end{tabular}
\end{table}

\begin{table}[ht]
\centering
\caption{\textbf{Benchmark results on VisA} for the Dinomaly model in the continual learning setting. Results are reported as: \textit{Image F1 | Pixel F1 | GFLOPs | Memory (MB)}.}
\label{tab:dinomaly_complete}
\footnotesize
\setlength{\tabcolsep}{3pt}
\begin{tabular}{ll cc}
\toprule
\textbf{AD Model} & \textbf{Strategy} & \textbf{DeiT Tiny} & \textbf{ViT Base} \\
\midrule
\multirow{2}{*}{Dinomaly} & cl 100 & 0.78 | 0.33 | 3.39 | 50.87 & 0.81 | 0.32 | 68.56 | 575.66 \\
                          & cl 40  & 0.77 | 0.33 | 3.39 | 42.25 & 0.86 | 0.38 | 68.56 | 567.04 \\
\bottomrule
\end{tabular}
\end{table}

\begin{figure}[htbp]
    \centering
    \includegraphics[width=0.75\textwidth]{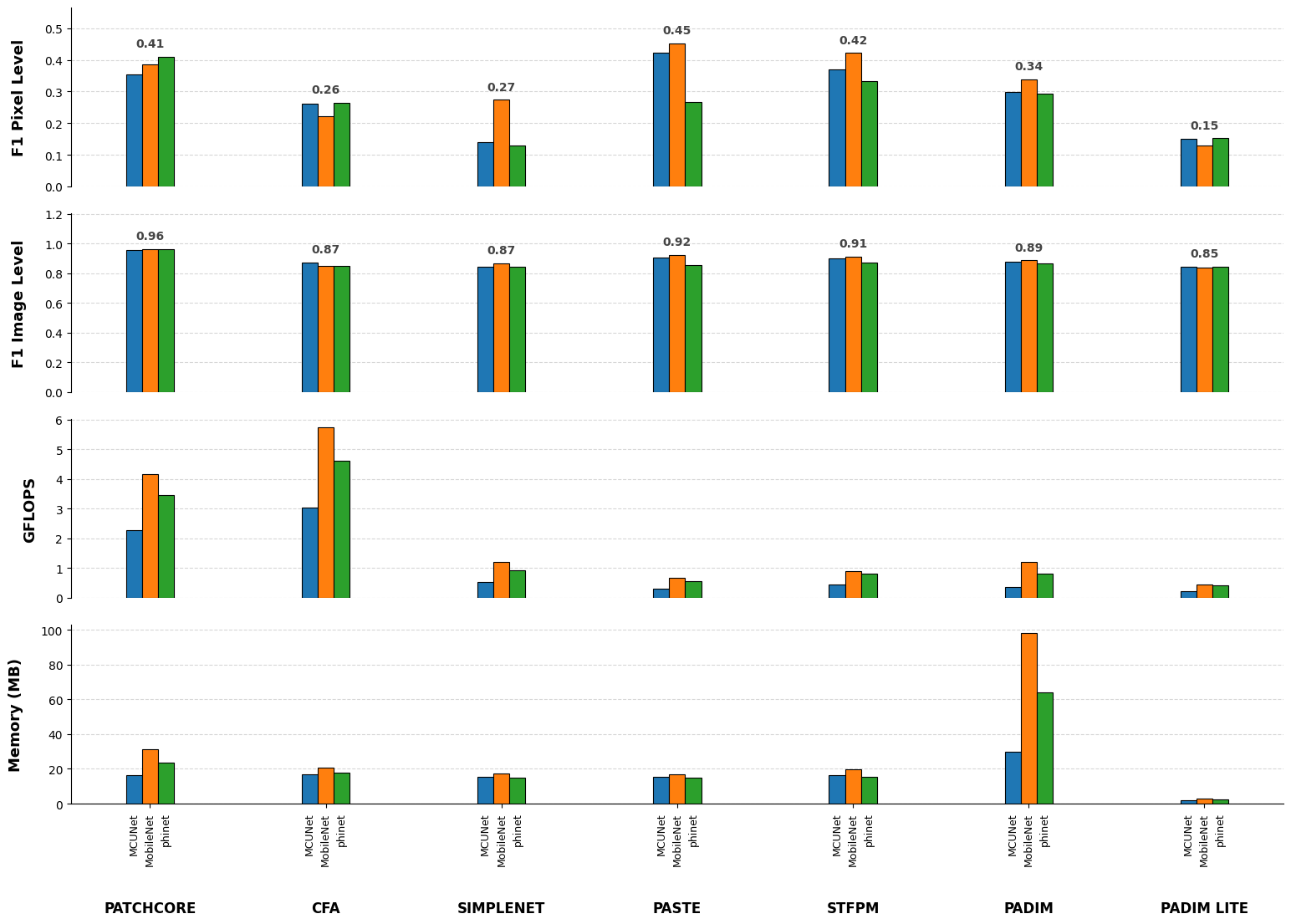}
    \caption{Evaluation of VAD models across different backbones on the MVTec Dataset. The charts compare F1 Pixel Level, F1 Image Level, computational complexity (GFLOPS), and Memory (MB). }
    \label{fig:grid_bar_plot_2}
\end{figure}

Figure~\ref{fig:cl_mobilenet_mvtec} illustrate the evolution of Pixel F1 across sequential tasks for all considered methods using MobileNet as backbone on MVTec AD.
Table \ref{tab:anomaly_results}  reports the full benchmark results across all model–backbone combinations, evaluated under the continual learning (CL) setting with replay buffer sizes of 40 and 100 samples, on the MVTec AD dataset. Results are expressed as  Image F1 / Pixel F1 / memory footprint in Megabytes (MB), and GFLOPs for inference cost.
\\
The benchmark reveals a clear tension between pixel-level localization quality, memory footprint, and inference cost, a tension that becomes especially acute in the edge continual learning setting, where all three dimensions are simultaneously constrained.
The first observation is that replacing standard backbones such as WideResNet50 with lightweight alternatives like MobileNet or MCUNet already yields substantial reductions in memory footprint and inference cost, often exceeding 85\%, with only marginal drops in detection performance.
While simple, this represents a significant step toward edge deployability for most VAD methods. 
\\
Nevertheless, targeted architectural design choices can push efficiency further while preserving performance.
Among all configurations evaluated, Tiny-Dinomaly and PaSTe emerge as the best-performing methods for pixel-level localization under edge constraints, combining competitive Pixel F1 scores with low inference costs and memory footprints suitable for resource-constrained devices. PatchCoreCL++ and PaDiM Lite MultiModal represent strong alternatives: while their pixel-level scores are slightly lower, both operate within acceptable computational and memory budgets, making them viable when image-level detection accuracy or distributional modeling are prioritized alongside localization.
\\
CFA and SimpleNet trail the field on pixel-level localization, with CFA peaking at 0.22 and SimpleNet at 0.27 with MobileNet. While both are computationally affordable, their Pixel F1 scores approach the lower bound established by fine-tuning baselines, suggesting they lack the spatial modeling capacity required for reliable anomaly localization in the continual learning setting.
\\
Considering all three dimensions jointly, no single method dominates across the board, and the optimal choice depends on the specific deployment constraints. PaSTe offers the most balanced configuration overall, combining competitive Pixel F1 with the lowest inference cost in the benchmark and a memory footprint below 16 MB.
When slightly higher memory is acceptable, Tiny-Dinomaly provides the best pixel-level localization at a still-affordable computational cost, making it the recommended choice when anomaly localization is the primary objective. For scenarios where image-level detection is paramount, PatchCoreCL++ delivers the highest image-level F1 at a fraction of the inference required by standard PatchCoreCL. PaDiM Lite MultiModal occupies a principled middle ground, trading a moderate memory overhead for localization quality that surpasses most other distribution-based alternatives within the family.

\subsection{Backbone Effect}

Backbone selection has a consistent impact on the efficiency-performance trade-off, with effects that generalize across most VAD methods evaluated in this benchmark (Fig.~\ref{fig:backbone_comparison}). 
MobileNet systematically yields the highest Pixel F1 scores within each model family, for instance, enabling PaSTe to reach 0.45, STFPM to 0.42, and PaDiM-Lite to 0.41, but does so at the cost of the highest memory consumption among the three lightweight backbones, as clearly visible in the Memory panel of Figure ~\ref{fig:backbone_comparison}. 
This makes it the recommended choice when localization quality is the primary objective and memory constraints allow for allocations in the 17-50 MB range, depending on the method.
\\
MCUNet offers the best inference-cost-to-performance ratio across most methods. It consistently delivers Pixel F1 scores within 3–5 percentage points of MobileNet while reducing GFLOPs substantially, for example, bringing PaSTe inference from 0.68 to 0.32 GFLOPs and PatchCoreCL++ from 4.15 to 2.29 GFLOPs, with memory footprints remaining comparable or slightly lower. This makes MCUNet the preferred backbone when latency is the main constraint, like in the case of real-time edge deployment.
\\
Eventually, PhiNet occupies an intermediate position: it achieves inference costs comparable to or lower than MobileNet, for instance, 0.81 GFLOPs vs. 0.90 for STFPM and 0.56 vs. 0.68 GFLOPs for PaSTe, while maintaining memory footprints between those of MCUNet and MobileNet. 


\subsection{Replay Memory Size Effect}
The replay memory size is relevant since it will influence the performance of models directly in the continual learning setting.
Based on the results, a buffer of 40 samples (\textasciitilde5 MB) already captures most of the available performance, demonstrating that even minimal replay budgets are sufficient to effectively 
mitigate catastrophic forgetting in VAD settings. This is a practically significant 
result: it implies that edge-deployable continual learning does not require large 
memory buffers, and that most VAD methods evaluated in this benchmark can be 
updated incrementally with a small additional storage overhead. Scaling to 100 
samples (\textasciitilde15 MB) yields consistent but moderate gains. 
Beyond this threshold, returns diminish rapidly: preliminary experiments with a buffer of 300 samples (~45 MB), not included in the main benchmark tables as they fall outside the realistic edge memory budgets targeted in this work, showed only a negligible improvement over the 100-sample configuration. This suggests that models saturate their effective generalization capacity at relatively modest buffer sizes, confirming that the replay buffer sizes of 40 and 100 adopted in our systematic evaluation are sufficient to characterize the performance-memory trade-off relevant to edge deployment.

\subsection{Tiny-Dinomaly}
At the image level, Tiny-Dinomaly achieves competitive performance, with an Image F1 of 0.88 and Pixel F1 of 0.45 using a replay memory size of 100, achieving a gap of 1 percentage point behind Dinomaly.
\\
More striking, however, is Tiny-Dinomaly's advantage at the pixel level, where it reaches a Pixel F1 of 0.45 compared to 0.40 for its larger counterpart. 
This gap is all the more remarkable given how much VAD models typically struggle to gain even a fraction of a percentage point on this metric.
As discussed above, this improvement stems from the stronger implicit regularization of the DeiT architecture, which constrains overfitting to the training distribution and yields better-calibrated anomaly maps. This effect is especially pronounced for pixel-level localization, where the quality of the spatial score distribution matters more than raw representational power.
\\
From a computational standpoint, DeiT offers a substantially more efficient alternative: it requires approximately 20× fewer GFLOPs in inference (3.39 vs. 68.56 GFLOPs), and its parameter footprint is reduced by a factor of 11x (50.87 MB vs. 575.66 MB). In resource-constrained deployment scenarios, such as edge computing or real-time industrial inspection, these efficiency gains represent a critical practical advantage, particularly when considered alongside Tiny-Dinomaly's superior anomaly localization performance.

\subsection{PadimCL and PadimCL-Lite}
We begin with a comparison between the original PaDiM-CL formulation proposed in \cite{bugarin2024unveiling} and our corrected version, with results reported in Table~\ref{tab:padimcl_comparison}.
As described in Section~\ref{sec:methodology}, the original PaDiM-CL implementation assigns uniform weights across tasks when computing the incremental mean and covariance, regardless of how many samples each task contributes. This introduces a systematic bias whenever tasks differ in dataset size. The impact is clearly reflected in the results: the original PaDiM-CL achieves an Image ROC-AUC of 0.56 and a Pixel F1 of 0.17, whereas our corrected formulation, PaDiM-CL UniModal, reaches 0.9 and 0.48 (see Table ~\ref{tab:padimcl_comparison}), respectively, a substantial improvement across all metrics with the WideResNet backbone.
\\
While PaDiM-CL UniModal represents a meaningful step forward, a substantial performance gap persists relative to PaDiM-CL MultiModal. PaDiM-CL UniModal using the MobileNet backbone achieves an Img F1 of 0.89 and a Pxl F1 of 0.34, whereas PaDiM-CL MultiModal reaches 0.94 and 0.5, respectively (see Table \ref{tab:anomaly_results}). This is caused by collapsing task-specific Gaussian distributions into a single merged distribution, which introduces a non-negligible approximation error, especially pronounced at the pixel level.
Yet this performance advantage comes at a steep cost, PaDiM-CL MultiModal maintains one full covariance matrix per task, causing memory to grow unboundedly with the number of tasks. With MobileNet, this amounts to 1,469.91 MB and 11.96 GFLOPs per sample, roughly 15× the 98.0 MB and 1.22 GFLOPs required by the unimodal variant, making it infeasible for edge deployment.
\\
\\
This tension between accuracy and resource consumption motivates the introduction of the lite variants, which replace full covariance matrices with diagonal approximations. 
However, PaDiM-CL-Lite UniModal has a significant drop not only in memory but also in performance, with Img F1 0.84 and Pxl F1 0.13, with values close to the lower bound of fine-tuning  (see Table \ref{tab:anomaly_results}).
\\
PaDiM-CL-Lite MultiModal represents a better alternative. While it also grows linearly like PaDiM-CL MultiModal, it grows only linearly in the number of features compared to the quadratic increase of PaDiM-CL MultiModal.
On MVTec with MobileNet, this yields an Img F1 of 0.9 and a Pxl F1 of 0.41, substantially above both PaDiM-CL UniModal, at a memory footprint of only 46.67 MB and an inference cost of 0.57 GFLOPs, compared to 98.0 MB and 1.22 GFLOPs for PaDiM-CL UniModal.
These results establish PaDiM-CL-Lite MultiModal as the most compelling trade-off in the PaDiM-CL family for edge-constrained continual learning.

\subsection{PatchCoreCL and PatchCoreCL++}
Compared to WideResNet50 with a memory bank of 30,000, PatchCore with MobileNet achieves a slight performance drop, Img F1 decreasing from 0.977 to 0.949 and Pxl F1 from 0.466 to 0.376, but with a significant reduction in model memory from 299.92 MB to 31.04 MB (approximately 90\%) and in inference cost from 357.77 GFLOPs to 49.16 GFLOPs (approximately 86\%). 

In addition, examining the impact of memory bank size, reducing the bank from 30,000 to 10,000 entries results in a moderate drop in Img F1 from 0.949 to 0.932 and in Pxl F1 from 0.376 to 0.335. This confirms that larger memory banks provide consistent, albeit moderate, gains across all metrics, at the cost of increased storage requirements.

When comparing the original PatchCoreCL and the proposed PatchCoreCL++, two observations stand out. First, PatchCoreCL++ yields a slight improvement in detection performance, with Img F1 increasing from 0.932 to 0.947 (+1.5 percentage points) and Pxl F1 from 0.335 to 0.358 (+2.3 percentage points) at 10k.
This improvement stems from the truncation heuristic used to prune the least representative samples from each memory bank: rather than rerunning the full k-center algorithm, simply discarding the tail of the coreset list not only reduces the memory bank update cost but also proves more effective than more elaborate recompression strategies. 

Moreover, this performance gain is accompanied by a substantial reduction in inference cost, thanks to the prototype-based memory bank selection. Specifically, inference cost drops from 49.16 GFLOPs for PatchCoreCL to 4.15 GFLOPs for PatchCoreCL++ at 30k, corresponding to a 12× reduction, making the proposed variant significantly more suitable for deployment on resource-constrained edge devices.

\section{Conclusion}
\label{sec:conclusion}

This work presents the first comprehensive benchmark for Visual Anomaly Detection under the combined constraints of edge deployment and continual learning. We characterized the trade-offs between detection performance, memory footprint, and inference cost in a setting that closely resembles real-world industrial requirements by testing seven VAD models across three lightweight backbone architectures on both MVTec AD and VisA.

Our results indicate that substituting lightweight backbones for conventional backbones already results in significant efficiency gains, with only slight decreases in detection performance.
In addition, replay buffer size and backbone selection are both significant variables. The majority of the continual learning performance is already captured by a buffer of 40 samples (5 MB), with only modest gains from scaling to 100 samples.
Eventually, no single method dominates across all three dimensions simultaneously, and the optimal configuration is inherently deployment-specific.

Beyond the benchmark, we introduced three novel contributions. 
Tiny-Dinomaly, achieves a 11× reduction in memory footprint and a 20× reduction in inference cost compared to its larger counterpart Dinomaly, while simultaneously improving Pixel F1 by 5 percentage points, a result that demonstrates how structural regularization from a compact encoder can outperform explicit architectural constraints designed for larger models.

PatchCoreCL++ addresses two practical limitations of the original PatchCoreCL: 
the cost of memory bank recompression and the linear growth of inference time with the number of tasks, reducing inference cost by 12×, making it substantially more practical for edge scenarios.

We corrected a systematic bias in the original PaDiM-CL UniModal formulation, where uniform task weighting during incremental Gaussian fusion introduced significant estimation errors when tasks differ in dataset size. Building on this, we introduced PaDiM-CL-Lite MultiModal, which replaces full covariance matrices with diagonal approximations while retaining a separate distribution per task. 
The method substantially outperforms PaDiM-CL UniModal reducing peak memory from 98 MB to just 46.67 MB
while achieving a Pixel F1 of 0.41 (0.34 in the original).
Padim-CL-Lite MultiModal emerges as the most favorable memory–accuracy trade-off in the PaDiM family for edge-constrained continual learning.

Taken together, these results suggest that efficient, adaptive VAD on tiny devices is feasible without sacrificing localization quality. 
Future work will investigate compression strategies for the replay memory to reduce the storage. Additionally, we will explore CL techniques specifically designed to operate within the strict memory and computational budgets imposed by edge hardware, moving toward light adaptation mechanisms.

{
    \small
    \bibliographystyle{plain}
    \bibliography{main}

@inproceedings{chaudhry2019continual,
  title={Continual learning with tiny episodic memories},
  author={Chaudhry, Arslan and Rohrbach, Marcus and Elhoseiny, Mohamed and Ajanthan, Thalaiyasingam and Dokania, P and Torr, P and Ranzato, M},
  booktitle={Workshop on Multi-Task and Lifelong Reinforcement Learning},
  year={2019}
}

@inproceedings{zhang2023destseg,
  title={Destseg: Segmentation guided denoising student-teacher for anomaly detection},
  author={Zhang, Xuan and Li, Shiyu and Li, Xi and Huang, Ping and Shan, Jiulong and Chen, Ting},
  booktitle={Proceedings of the IEEE/CVF conference on computer vision and pattern recognition},
  pages={3914--3923},
  year={2023}
}

@InProceedings{Deng_2022_CVPR,
author    = {Deng, Hanqiu and Li, Xingyu},
title     = {Anomaly Detection via Reverse Distillation From One-Class Embedding},
booktitle = {Proceedings of the IEEE/CVF Conference on Computer Vision and Pattern Recognition (CVPR)},
month     = {June},
year      = {2022},
pages     = {9737-9746}}

@article{zavrtanik2021reconstruction,
  title={Reconstruction by inpainting for visual anomaly detection},
  author={Zavrtanik, Vitjan and Kristan, Matej and Sko{\v{c}}aj, Danijel},
  journal={Pattern Recognition},
  volume={112},
  pages={107706},
  year={2021},
  publisher={Elsevier}
}

@inproceedings{sandler2018mobilenetv2,
  title={Mobilenetv2: Inverted residuals and linear bottlenecks},
  author={Sandler, Mark and Howard, Andrew and Zhu, Menglong and Zhmoginov, Andrey and Chen, Liang-Chieh},
  booktitle={Proceedings of the IEEE conference on computer vision and pattern recognition},
  pages={4510--4520},
  year={2018}
}

@article{lin2020mcunet,
  title={Mcunet: Tiny deep learning on iot devices},
  author={Lin, Ji and Chen, Wei-Ming and Lin, Yujun and Gan, Chuang and Han, Song and others},
  journal={Advances in neural information processing systems},
  volume={33},
  pages={11711--11722},
  year={2020}
}

@article{phinet,
author = {Paissan, Francesco and Ancilotto, Alberto and Farella, Elisabetta},
title = {PhiNets: A Scalable Backbone for Low-power AI at the Edge},
year = {2022},
issue_date = {September 2022},
publisher = {Association for Computing Machinery},
address = {New York, NY, USA},
volume = {21},
number = {5},
issn = {1539-9087},
url = {https://doi.org/10.1145/3510832},
doi = {10.1145/3510832},
journal = {ACM Trans. Embed. Comput. Syst.},
month = {dec},
articleno = {53},
numpages = {18}
}

@article{zagoruyko2016wide,
  title={Wide residual networks},
  author={Zagoruyko, Sergey and Komodakis, Nikos},
  journal={arXiv preprint arXiv:1605.07146},
  year={2016}
}

@inproceedings{bao2024bmad,
  title={Bmad: Benchmarks for medical anomaly detection},
  author={Bao, Jinan and Sun, Hanshi and Deng, Hanqiu and He, Yinsheng and Zhang, Zhaoxiang and Li, Xingyu},
  booktitle={Proceedings of the IEEE/CVF conference on computer vision and pattern recognition},
  pages={4042--4053},
  year={2024}
}

@inproceedings{he2024diffusion,
  title={A diffusion-based framework for multi-class anomaly detection},
  author={He, Haoyang and Zhang, Jiangning and Chen, Hongxu and Chen, Xuhai and Li, Zhishan and Chen, Xu and Wang, Yabiao and Wang, Chengjie and Xie, Lei},
  booktitle={Proceedings of the AAAI conference on artificial intelligence},
  volume={38},
  number={8},
  pages={8472--8480},
  year={2024}
}

@article{zhang2025diffusionad,
  title={DiffusionAD: Norm-guided one-step denoising diffusion for anomaly detection},
  author={Zhang, Hui and Wang, Zheng and Zeng, Dan and Wu, Zuxuan and Jiang, Yu-Gang},
  journal={IEEE transactions on pattern analysis and machine intelligence},
  year={2025},
  publisher={IEEE}
}

@article{huang2024prototype,
  title={A prototype-based neural network for image anomaly detection and localization},
  author={Huang, Chao and Kang, Zhao and Wu, Hong},
  journal={Neural Processing Letters},
  volume={56},
  number={3},
  pages={169},
  year={2024},
  publisher={Springer}
}

@inproceedings{zavrtanik2021draem,
  title={Draem-a discriminatively trained reconstruction embedding for surface anomaly detection},
  author={Zavrtanik, Vitjan and Kristan, Matej and Sko{\v{c}}aj, Danijel},
  booktitle={Proceedings of the IEEE/CVF international conference on computer vision},
  pages={8330--8339},
  year={2021}
}

@article{yu2021fastflow,
  title={Fastflow: Unsupervised anomaly detection and localization via 2d normalizing flows},
  author={Yu, Jiawei and Zheng, Ye and Wang, Xiang and Li, Wei and Wu, Yushuang and Zhao, Rui and Wu, Liwei},
  journal={arXiv preprint arXiv:2111.07677},
  year={2021}
}

@inproceedings{gudovskiy2022cflow,
  title={Cflow-ad: Real-time unsupervised anomaly detection with localization via conditional normalizing flows},
  author={Gudovskiy, Denis and Ishizaka, Shun and Kozuka, Kazuki},
  booktitle={Proceedings of the IEEE/CVF winter conference on applications of computer vision},
  pages={98--107},
  year={2022}
}

@inproceedings{caron2021emerging,
  title={Emerging properties in self-supervised vision transformers},
  author={Caron, Mathilde and Touvron, Hugo and Misra, Ishan and J{\'e}gou, Herv{\'e} and Mairal, Julien and Bojanowski, Piotr and Joulin, Armand},
  booktitle={Proceedings of the IEEE/CVF international conference on computer vision},
  pages={9650--9660},
  year={2021}
}

@inproceedings{defard2021padim,
  title         = {Padim: a patch distribution modeling framework for anomaly detection and localization},
  author        = {Defard, Thomas and Setkov, Aleksandr and Loesch, Angelique and Audigier, Romaric},
  booktitle     = {International conference on pattern recognition},
  pages         = {475--489},
  year          = {2021},
  organization  = {Springer}
}

@inproceedings{roth2022towards,
  title         = {Towards total recall in industrial anomaly detection},
  author        = {Roth, Karsten and Pemula, Latha and Zepeda, Joaquin and Sch{\"o}lkopf, Bernhard and Brox, Thomas and Gehler, Peter},
  booktitle     = {Proceedings of the IEEE/CVF conference on computer vision and pattern recognition},
  pages         = {14318--14328},
  year          = {2022}
}

@misc{wang2021studentteacherfeaturepyramidmatching,
  title         = {Student-Teacher Feature Pyramid Matching for Anomaly Detection},
  author        = {Guodong Wang and Shumin Han and Errui Ding and Di Huang},
  year          = {2021},
  eprint        = {2103.04257},
  archiveprefix = {arXiv},
  primaryclass  = {cs.CV},
  url           = {https://arxiv.org/abs/2103.04257}
}

@article{9839549,
  author        = {Lee, Sungwook and Lee, Seunghyun and Song, Byung Cheol},
  journal       = {IEEE Access},
  title         = {CFA: Coupled-Hypersphere-Based Feature Adaptation for Target-Oriented Anomaly Localization},
  year          = {2022},
  volume        = {10},
  number        = {},
  pages         = {78446--78454},
  doi           = {10.1109/ACCESS.2022.3193699}
}

@inproceedings{liu2023simplenet,
  title={Simplenet: A simple network for image anomaly detection and localization},
  author={Liu, Zhikang and Zhou, Yiming and Xu, Yuansheng and Wang, Zilei},
  booktitle={Proceedings of the IEEE/CVF conference on computer vision and pattern recognition},
  pages={20402--20411},
  year={2023}
}

@inproceedings{guo2025dinomaly,
  title={Dinomaly: The less is more philosophy in multi-class unsupervised anomaly detection},
  author={Guo, Jia and Lu, Shuai and Zhang, Weihang and Chen, Fang and Li, Huiqi and Liao, Hongen},
  booktitle={Proceedings of the Computer Vision and Pattern Recognition Conference},
  pages={20405--20415},
  year={2025}
}

@inproceedings{barusco2025paste,
  title         = {Paste: Improving the efficiency of visual anomaly detection at the edge},
  author        = {Barusco, Manuel and Borsatti, Francesco and Pezze, Davide Dalle and Paissan, Francesco and Farella, Elisabetta and Susto, Gian Antonio},
  booktitle     = {Proceedings of the Computer Vision and Pattern Recognition Conference},
  pages         = {4026--4035},
  year          = {2025}
}

@article{stropeni2026efficient,
  title={Efficient Visual Anomaly Detection at the Edge: Enabling Real-Time Industrial Inspection on Resource-Constrained Devices},
  author={Stropeni, Arianna and Genilotti, Fabrizio and Borsatti, Francesco and Barusco, Manuel and Pezze, Davide Dalle and Susto, Gian Antonio},
  journal={arXiv preprint arXiv:2603.20288},
  year={2026}
}

@article{barusco2025memory,
  title         = {Memory Efficient Continual Learning for Edge-Based Visual Anomaly Detection},
  author        = {Barusco, Manuel and D'Antoni, Lorenzo and Borsatti, Francesco and Dalle Pezze, Davide and Susto, Gian Antonio},
  journal       = {IFAC-PapersOnLine},
  volume        = {59},
  number        = {26},
  pages         = {85--90},
  year          = {2025},
  publisher     = {Elsevier}
}

@inproceedings{bugarin2024unveiling,
  title={Unveiling the anomalies in an ever-changing world: A benchmark for pixel-level anomaly detection in continual learning},
  author={Bugarin, Nikola and Bugaric, Jovana and Barusco, Manuel and Pezze, Davide Dalle and Susto, Gian Antonio},
  booktitle={Proceedings of the IEEE/CVF conference on computer vision and pattern recognition},
  pages={4065--4074},
  year={2024}
}

@inproceedings{li2025one,
  title={One-for-more: Continual diffusion model for anomaly detection},
  author={Li, Xiaofan and Tan, Xin and Chen, Zhuo and Zhang, Zhizhong and Zhang, Ruixin and Guo, Rizen and Jiang, Guanna and Chen, Yulong and Qu, Yanyun and Ma, Lizhuang and others},
  booktitle={Proceedings of the Computer Vision and Pattern Recognition Conference},
  pages={4766--4775},
  year={2025}
}

@inproceedings{li2022towards,
  title={Towards continual adaptation in industrial anomaly detection},
  author={Li, Wujin and Zhan, Jiawei and Wang, Jinbao and Xia, Bizhong and Gao, Bin-Bin and Liu, Jun and Wang, Chengjie and Zheng, Feng},
  booktitle={Proceedings of the 30th ACM International Conference on Multimedia},
  pages={2871--2880},
  year={2022}
}

@inproceedings{liu2024unsupervised,
  title={Unsupervised continual anomaly detection with contrastively-learned prompt},
  author={Liu, Jiaqi and Wu, Kai and Nie, Qiang and Chen, Ying and Gao, Bin-Bin and Liu, Yong and Wang, Jinbao and Wang, Chengjie and Zheng, Feng},
  booktitle={Proceedings of the AAAI conference on artificial intelligence},
  volume={38},
  number={4},
  pages={3639--3647},
  year={2024}
}

@inproceedings{touvron2021training,
  title={Training data-efficient image transformers \& distillation through attention},
  author={Touvron, Hugo and Cord, Matthieu and Douze, Matthijs and Massa, Francisco and Sablayrolles, Alexandre and J{\'e}gou, Herv{\'e}},
  booktitle={International conference on machine learning},
  pages={10347--10357},
  year={2021},
  organization={PMLR}
}

@article{oquab2023dinov2,
  title={Dinov2: Learning robust visual features without supervision},
  author={Oquab, Maxime and Darcet, Timoth{\'e}e and Moutakanni, Th{\'e}o and Vo, Huy and Szafraniec, Marc and Khalidov, Vasil and Fernandez, Pierre and Haziza, Daniel and Massa, Francisco and El-Nouby, Alaaeldin and others},
  journal={arXiv preprint arXiv:2304.07193},
  year={2023}
}

@article{barusco2025moviad,
  title         = {Moviad: A modular library for visual anomaly detection},
  author        = {Barusco, Manuel and Borsatti, Francesco and Stropeni, Arianna and Pezze, Davide Dalle and Susto, Gian Antonio},
  journal       = {arXiv preprint arXiv:2507.12049},
  year          = {2025}
}

@inproceedings{bergmann2019mvtec,
  title         = {MVTec AD--A comprehensive real-world dataset for unsupervised anomaly detection},
  author        = {Bergmann, Paul and Fauser, Michael and Sattlegger, David and Steger, Carsten},
  booktitle     = {Proceedings of the IEEE/CVF conference on computer vision and pattern recognition},
  pages         = {9592--9600},
  year          = {2019}
}

@inproceedings{zou2022spot,
  title={Spot-the-difference self-supervised pre-training for anomaly detection and segmentation},
  author={Zou, Yang and Jeong, Jongheon and Pemula, Latha and Zhang, Dongqing and Dabeer, Onkar},
  booktitle={European conference on computer vision},
  pages={392--408},
  year={2022},
  organization={Springer}
}

@article{gonzalez1985clustering,
  title={Clustering to minimize the maximum intercluster distance},
  author={Gonzalez, Teofilo F},
  journal={Theoretical computer science},
  volume={38},
  pages={293--306},
  year={1985},
  publisher={Elsevier}
}
}

\end{document}